\documentclass[11pt, a4paper, logo, copyright, nonumbering]{map}
\usepackage{CJKutf8}
\usepackage{xargs}  
\usepackage{todonotes}
\usepackage{amsmath}
\usepackage{dsfont}
\usepackage{algorithmicx}

\usepackage{longtable}

\usepackage{svg}
\usepackage{fontawesome}
\usepackage{array}
\usepackage{tabularx}
\usepackage{latexsym}
\usepackage{graphicx}
\usepackage[utf8]{inputenc} 
\usepackage{amssymb} 
\usepackage{url}            
\usepackage{amsfonts}       
\usepackage{nicefrac}       
\usepackage{microtype}      
\usepackage{xspace}

\usepackage{listings}
\usepackage{makecell}
\usepackage{inconsolata}
\usepackage{multicol}
\usepackage{amstext}

\definecolor{darkblue}{RGB}{84, 112, 198}

\definecolor{lightblue}{rgb}{0.85, 0.95, 1.0}    
\definecolor{lightgreen}{rgb}{0.90, 1.0, 0.90}    
\definecolor{lightorange}{rgb}{1.0, 0.95, 0.85}   
\definecolor{lightpurple}{rgb}{0.95, 0.90, 1.0}   
\definecolor{lightgray}{rgb}{0.97, 0.97, 0.97}    

\usepackage{tikz}
\usetikzlibrary{calc}
\definecolor{battery-empty}{rgb}{0.9, 0.9, 0.9}
\newcommand{\difficultybar}[1]{%
  \begin{tikzpicture}[baseline, scale=0.5, every node/.style={scale=0.8}]
    \foreach \i in {1,2,3,4,5} {
      \ifnum\i>#1
        \draw[fill=battery-empty] (\i*0.5-0.5, 0) rectangle (\i*0.5, 0.25);
      \else
        \pgfmathsetmacro{\colorlevel}{80 - 12*(\i)} 
        \edef\x{\noexpand\draw[fill=blue!\colorlevel!white, opacity=0.9] (\i*0.5-0.5, 0) rectangle (\i*0.5, 0.25);}
        \x
        \draw[blue!50!black] (\i*0.5-0.5, 0) rectangle (\i*0.5, 0.25);
      \fi
    }
    \fill[battery-empty!70] (2.5, 0.08) rectangle (2.6, 0.17);
    \draw[battery-empty!70!black] (2.5, 0.08) rectangle (2.6, 0.17);
  \end{tikzpicture}%
}

\definecolor{hidden-draw}{RGB}{20,68,106}
\definecolor{hidden-pink}{RGB}{255,245,247}
\usepackage[edges]{forest}

\usepackage{bm}

\usepackage{natbib}
\usepackage{CJKutf8}
\usepackage{xargs}
\usepackage{todonotes}
\usepackage{multirow}
\usepackage{cleveref}
\usepackage{amsmath}
\usepackage{dsfont}
\usepackage{subcaption}
\usepackage{caption}
\usepackage{url}
\usepackage{svg}
\usepackage{fontawesome}
\usepackage{xcolor}
\usepackage{float}

\usepackage[utf8]{inputenc}
\usepackage{booktabs}
\usepackage{graphicx}
\usepackage{array}
\usepackage{tabularx}
\usepackage{xcolor,colortbl}  
\definecolor{boxcolor}{HTML}{d92523} 
\definecolor{bulbcolor}{HTML}{e3b87f} 
\usepackage{url}
\usepackage{ragged2e}   
\usepackage{makecell} 

\usepackage{hyperref}       
\usepackage{url}            
\usepackage{booktabs}       
\usepackage{amsfonts}       
\usepackage{nicefrac}       
\usepackage{microtype}      
\usepackage{xcolor}         
\usepackage{graphicx}
\usepackage{longtable}
\usepackage{array}
\usepackage{pbox}
\usepackage{amsmath}
\usepackage{amssymb}
\usepackage{tabularx}
\usepackage{multirow}
\usepackage{wrapfig}
\usepackage{pifont}
\usepackage{algorithm}
\usepackage{algpseudocode}
\usepackage{lipsum}
\usepackage{subcaption}
\usepackage{enumitem}
\usepackage{tikz}
\usepackage{xstring}

\usepackage{color-edits}

\usepackage{booktabs}
\usepackage{multirow}
\usepackage[table]{xcolor}
\usepackage{tabularx}

\usepackage{parskip} 

\usepackage{threeparttable} 

\definecolor{rliableolive}{HTML}{BBCC33}
\definecolor{rliableblue}{HTML}{77AADD}
\definecolor{rliablered}{HTML}{f63c44}

\definecolor{rliableolive}{HTML}{BBCC33}
\definecolor{rliableblue}{HTML}{77AADD}
\definecolor{rliablered}{HTML}{f63c44}

\usepackage[most,skins,theorems]{tcolorbox}
\tcbuselibrary{skins,breakable,fitting,hooks}  
\tcbset{
  aibox/.style={
    width=\linewidth,
    top=7pt,
    bottom=2pt,
    colback=rliablered!18!white,
    colframe=black,
    colbacktitle=black,
    enhanced,
    center,
    attach boxed title to top left={yshift=-0.1in,xshift=0.15in},
    boxed title style={boxrule=0pt,colframe=white,},
  }
}
\tcbset{
  aibox2/.style={
    width=\linewidth,
    top=7pt,
    bottom=2pt,
    colback=green!18!white,
    colframe=black,
    colbacktitle=black,
    enhanced,
    center,
    attach boxed title to top left={yshift=-0.1in,xshift=0.15in},
    boxed title style={boxrule=0pt,colframe=white,},
  }
}
\newtcolorbox{AIbox}[2][]{aibox,title=#2,#1}
\newtcolorbox{AIbox2}[2][]{aibox2,title=#2,#1}

\hypersetup{
    colorlinks=true,            
    linkcolor=blue,             
    filecolor=magenta,          
    urlcolor=cyan,              
    citecolor=purple,             
    pdftitle={Overleaf Example},
    pdfpagemode=FullScreen,
}

\definecolor{iquestblue}{HTML}{173C7F}
\definecolor{iquestazure}{HTML}{528FCC}

\renewcommand{\today}{}
\newcommandx{\info}[2][1=]{\todo[linecolor=red,backgroundcolor=red!25,bordercolor=red,#1]{#2}}

\title{
TMAS: Scaling Test-Time Compute via Multi-Agent Synergy
}

\author{
George Wu$^{1*}$,
Nan Jing$^{1*}$,
Qing Yi$^{1*}$, \\
\textnormal{Chuan Hao$^{1\dagger}$,
Ming Yang$^{1}$,
Feng Chang$^{1}$,
Yuan Wei$^{1}$,
Jian Yang$^{2}$,
Ran Tao$^{1}$,
Bryan Dai$^{1}$}
\\
\textnormal{\small $^{1}$IQuest Research}\\
\textnormal{\small $^{2}$Beihang University}\\
\textnormal{\small $^*$Equal Contribution, $^{\dagger}$Corresponding Author}\\
\textnormal{\small \texttt{\{georgewzy01,yuxavierh,evanevanyang620\}@gmail.com,  chao@iquestlab.com}}
\vspace{-15pt} 
}

\begin{abstract}
\vspace{-0.9em}
Test-time scaling has become an effective paradigm for improving the reasoning ability of large language models by allocating additional computation during inference. Recent structured approaches have further advanced this paradigm by organizing inference across multiple trajectories, refinement rounds, and verification-based feedback. However, existing structured test-time scaling methods either weakly coordinate parallel reasoning trajectories or rely on noisy historical information without explicitly deciding what should be retained and reused, limiting their ability to balance exploration and exploitation. In this work, we propose TMAS, a framework for scaling test-time compute via multi-agent synergy. TMAS organizes inference as a collaborative process among specialized agents, enabling structured information flow across agents, trajectories, and refinement iterations. To support effective cross-trajectory collaboration, TMAS introduces hierarchical memories: the experience bank reuses low-level reliable intermediate conclusions and local feedback, while the guideline bank records previously explored high-level strategies to steer subsequent rollouts away from redundant reasoning patterns. Furthermore, we design a hybrid reward reinforcement learning scheme tailored to TMAS, which jointly preserves basic reasoning capability, enhances experience utilization, and encourages exploration beyond previously attempted solution strategies. Extensive experiments on challenging reasoning benchmarks show that TMAS achieves stronger iterative scaling than existing test-time scaling baselines, with hybrid reward training further improving scaling effectiveness and stability across iterations. On IMO-AnswerBench, TMAS improves the Pass@1 result of Qwen3-30B-A3B-Thinking-2507 from 64.9 to 77.2, while TMAS with Hybrid-RL improves Qwen3-4B-Thinking-2507 from 53.5 to 73.5, substantially narrowing the gap between small open-source backbones and much larger frontier models. Code and data are available at \url{https://github.com/IQuestLab/tmas}.
\end{abstract}
\vspace{-10pt}
\begin{document}

\maketitle

\let\oldthefootnote\thefootnote

\begin{figure}[!t]
    \centering
    \includegraphics[width=0.98\textwidth]{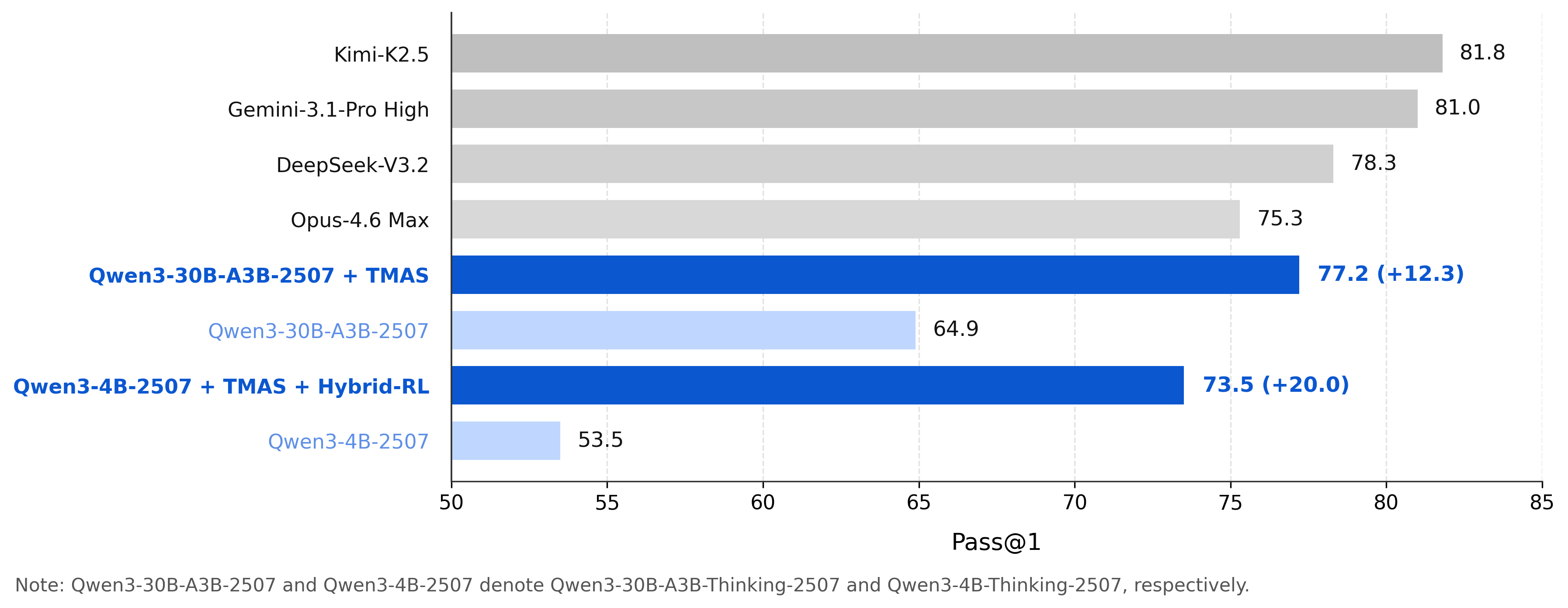}
    \caption{
Small backbones approach frontier models on IMO-AnswerBench. 
We compare TMAS-enhanced Qwen3 backbones with reported results from frontier models. 
Our enhanced systems are highlighted in blue, while the corresponding raw backbones are shown in light blue. 
TMAS improves Qwen3-30B-A3B-Thinking-2507 from 64.9\% to 77.2\% Pass@1, and TMAS with Hybrid-RL improves Qwen3-4B-Thinking-2507 from 53.5\% to 73.5\%, substantially narrowing the gap to much larger frontier models.
}
    \label{fig:full_imo_answerbench}
\end{figure}

\section{Introduction}

Test-time scaling (TTS) has emerged as an effective paradigm for improving the reasoning ability of large language models (LLMs) by allocating additional computation during inference. Early approaches mainly scale computation within a single generation, encouraging models to produce longer chains of thought or more deliberate reasoning processes~\cite{chain-of-thought,muennighoff2025s1, zhang2025alphaone}. As task difficulty increases, however, single-trajectory scaling becomes insufficient, motivating sequential and parallel forms of TTS that extend reasoning across multiple refinement rounds or multiple candidate trajectories~\citep{self-refine, self-consistency}. This evolution shifts the focus of TTS from merely increasing computation to more effectively organizing how reasoning trajectories are generated, refined, and reused.

Recent work has therefore explored structured hybrid architectures that jointly scale breadth and depth for difficult reasoning problems. One representative direction, including PaCoRe~\cite{pacore} and RSE~\cite{rse}, emphasizes inter-trajectory interaction by aggregating information from multiple historical attempts to guide subsequent reasoning. Another line of work adopts structured verify--refine paradigms, as in DeepSeek-Math-V2~\cite{deepseekmath-v2} and Nemotron-Cascade 2~\cite{nemotron-v2}, where multiple candidate solutions are generated and verified in parallel, followed by refinement based on explicit feedback. These systems can be naturally viewed through a multi-agent lens, with specialized components responsible for solution generation, verification, and refinement, interacting to progressively improve solution quality. Despite these advances, existing structured TTS methods still provide only limited collaboration among reasoning trajectories. Trajectory-aggregation methods improve inter-trajectory interaction, but they typically rely on large amounts of historical information without explicitly deciding what should be retained or discarded. Verify--refine systems introduce explicit feedback, but different trajectories are often weakly coupled, leaving useful findings and reusable experience insufficiently shared across attempts. Consequently, current methods either underutilize cross-trajectory experience or become overly constrained by noisy historical signals, limiting both exploration and exploitation. 

To address these limitations, we aim to extend existing multi-agent and parallel TTS paradigms with explicit cross-trajectory collaboration, where agents can extract, maintain, and propagate shared memory across reasoning trajectories. However, realizing such a framework requires addressing three key challenges. \textit{(1) Multi-agent synergy.}
A multi-agent TTS system must coordinate specialized agents within each trajectory while managing information flow across parallel trajectories and iterations. Without an explicit synergy mechanism, agent outputs may remain weakly aligned, and useful experience from one trajectory may fail to benefit others. Thus, an effective framework should define not only agent roles, but also how their outputs are organized, transmitted, and converted into reusable reasoning signals. \textit{(2) Hierarchical memory management.}
Memory is essential for long-horizon agentic reasoning, where multi-round interactions require persistent information to be retained and reused across iterations~\cite{hong2025context-rot,li2025-Mem-OS}. For complex problem solving, such memory must preserve both global solution strategies and reliable local reasoning states, such as verified anchors and intermediate conclusions. These signals differ in granularity and usage, yet existing methods often fail to distinguish them, limiting effective information sharing and reuse. \textit{(3) Exploration--exploitation balance.}
Solving difficult problems requires both exploring diverse hypotheses and exploiting accumulated evidence to refine promising directions~\cite{march1991exploration-exploitation,sutton1998reinforcement}. Similarly, test-time reasoning must explore diverse solution paths while exploiting reliable intermediate conclusions and accumulated experience. Without explicit control over this trade-off, models may either become trapped in suboptimal patterns or waste computation on redundant attempts.

Building on these observations, we propose \textbf{TMAS}, a framework for scaling \textbf{T}est-time compute via \textbf{M}ulti-\textbf{A}gent \textbf{S}ynergy. TMAS organizes test-time compute as a collaborative process among specialized agents, enabling structured information flow across agents, trajectories, and iterations. To address hierarchical memory management, TMAS introduces an experience agent and a guideline agent: the former maintains low-level experience memory, including concrete skills, local feedback, and reliable intermediate conclusions, while the latter records previously explored high-level strategies and structural insights to guide subsequent rollouts away from redundant solution patterns. To better align the model with TMAS, we further design a hybrid reward system consisting of three complementary training objectives: maintaining basic reasoning capability, enhancing experience utilization, and promoting exploration beyond previously attempted strategies. Together, these mechanisms strengthen the iterative scaling ability of TMAS, allowing additional test-time compute to be more effectively translated into improved performance on challenging reasoning problems. 

Our main contributions are summarized as follows:
\begin{itemize}[leftmargin=10pt, itemsep=0pt, parsep=3pt, topsep=3pt]
    \item We propose TMAS, a framework for scaling test-time compute via multi-agent synergy. TMAS explicitly organizes the flow of information across agents, trajectories, and iterations, transforming independent reasoning attempts into a coordinated iterative process. In particular, TMAS introduces experience and guideline agents to separately maintain low-level experience memory and high-level guideline memory, preserving reusable local reasoning signals while recording explored strategies to discourage redundancy and encourage diverse exploration.
    \item We design a hybrid reward RL scheme tailored to TMAS. Rather than optimizing only for final correctness, our training objective consists of three complementary tasks: preserving basic reasoning competence, enhancing experience utilization, and encouraging exploration beyond previously attempted strategies. This design enables the model to better exploit the collaborative memory structure of TMAS while maintaining sufficient exploration during iterative refinement.
    \item We conduct extensive experiments on challenging reasoning benchmarks. Results show that TMAS achieves stronger iterative scaling than existing TTS baselines, while hybrid reward RL further improves scaling effectiveness and stability across refinement rounds.
\end{itemize}

\section{Related Work}
\subsection{Test-Time Scaling}

Test-time scaling (TTS) enhances reasoning by allocating additional inference computation. Early paradigms mainly employ sequential scaling, such as Chain-of-Thought~\cite{chain-of-thought,qwen2,qwq-32b-preview} and Self-Refine~\cite{self-refine}, to extend or iteratively refine reasoning trajectories, or parallel scaling, such as Self-Consistency~\cite{self-consistency}, to aggregate independent solutions for error reduction. Search-based methods further structure this process through state expansion, evaluation, and pruning, as in Tree of Thoughts~\cite{tree-of-thought} and MCTS-based reasoning~\cite{hao2023reasoning,zhang2024rest}. Recent work has explored structured hybrid architectures that jointly scale breadth and depth for difficult reasoning problems. One line of work emphasizes inter-trajectory interaction and experience reuse: PaCoRe~\cite{pacore} synthesizes compact messages from parallel trajectories to guide subsequent rounds, while RSE~\cite{rse} distills historical trajectories into a shared experience bank. Another line adopts structured verify--refine paradigms~\cite{veri-refine}, where multiple candidate solutions are generated and verified in parallel, followed by refinement based on explicit feedback, DeepSeek-Math-V2~\cite{deepseekmath-v2}, Nemotron-Cascade 2~\cite{nemotron-v2}, and Alethia~\cite{alethia}. 
These methods can be viewed through a multi-agent lens, with specialized components for generation, verification, and refinement. However, existing TTS methods still lack effective collaboration across reasoning trajectories. Verify–refine frameworks introduce explicit feedback, yet reusable experience is often insufficiently shared across attempts. Trajectory-aggregation approaches improve inter-trajectory interaction, but typically accumulate historical information without explicitly selecting what should be retained, abstracted, or discarded, making them vulnerable to noisy or suboptimal signals. To address this limitation, TMAS explicitly organizes information flow across agents, trajectories, and iterations while introducing specialized memory agents to selectively maintain and reuse critical reasoning signals, improving the balance between experience exploitation and novel strategy exploration.

\subsection{Multi-Agent Systems for Mathematical Reasoning}

Multi-agent systems decompose mathematical reasoning into interacting roles. Early training-free, debate-style protocols utilizing frozen models~\cite{du2024improving,liang2024encouraging,zhang2025debate4math} often struggle with exceptionally challenging problems. Recent approaches introduce structured role decomposition to tackle harder tasks~\cite{veri-refine,luo2025learning,singh2026v_1}, yet still primarily rely on unadapted, frozen models. To bridge this gap, subsequent research~\cite{liu2025marsrl,zhang2026seed-scaling,chen2025magicore,alphaproof,seedprover} explicitly trains models for collaborative roles. For instance, MarsRL~\cite{liu2025marsrl} optimizes a solver--verifier--corrector pipeline via reinforcement learning (RL) with agent-specific rewards, demonstrating that effective multi-agent reasoning requires targeted training alongside structural design. Inspired by this progression, we introduce a lightweight hybrid reward system tailored for the TMAS framework. Our reward design preserves foundational reasoning capabilities while incentivizing experience utilization and novel strategy exploration, thereby enabling TMAS to optimally coordinate exploration and exploitation during iterative reasoning.

\section{Methods}
\label{sec:method}
\begin{figure}[!t]
    \centering
    \includegraphics[width=0.98\textwidth]{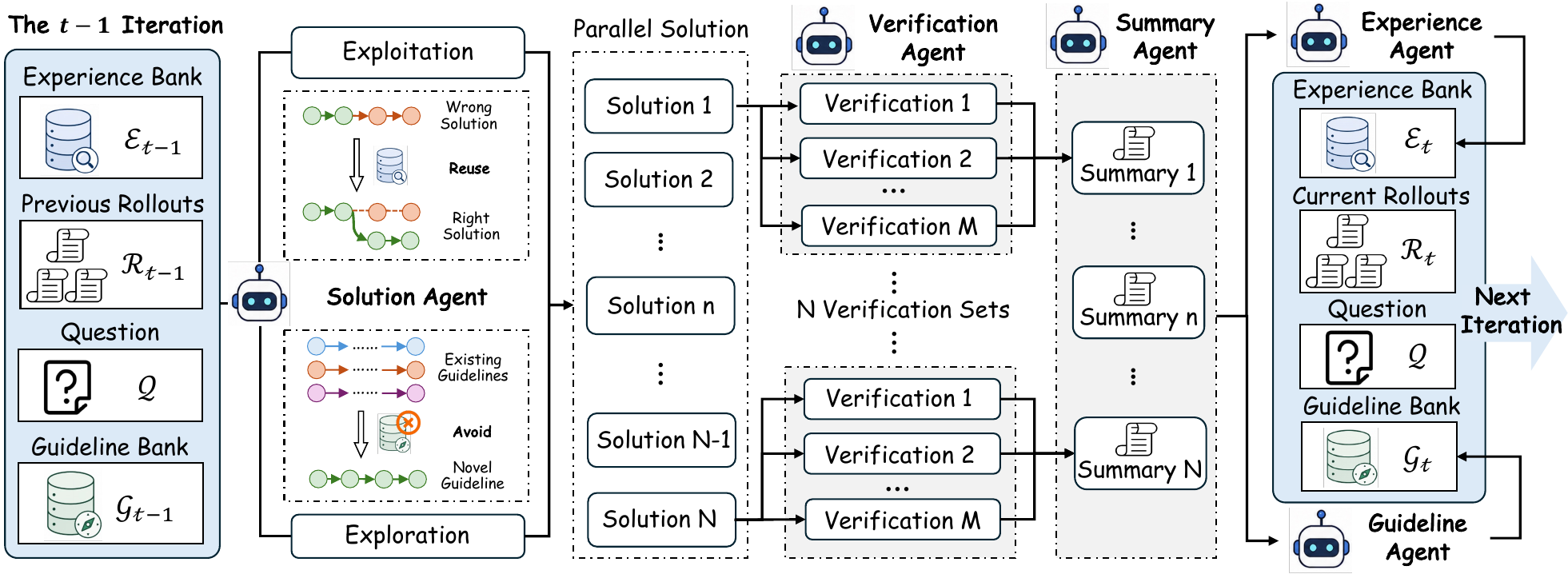}
    \caption{Overview of the TMAS framework. For each problem, TMAS generates multiple solution trajectories in parallel, verifies each trajectory with independent verifier agents, and summarizes the feedback into rollout-level summaries. The experience agent extracts reusable low-level reasoning signals into the experience bank, while the guideline agent records previously explored high-level strategies in the guideline bank. These two memory banks are then used in subsequent iterations to support experience-based refinement and non-redundant exploration.}
    \label{fig:overall_framework}
\end{figure}
\subsection{Overall Framework}
As illustrated in Figure~\ref{fig:overall_framework}, we propose TMAS, a framework for scaling test-time compute via multi-agent synergy, which integrates parallel exploration with sequential exploitation. At each iteration, TMAS explores multiple reasoning paths in parallel and accumulates useful signals from these paths for subsequent refinement. To organize this process, TMAS assigns five specialized agents to complementary functions, including solution generation, verification, summarization, and memory update. A memory-bank-based communication mechanism then coordinates these agents across parallel trajectories and refinement iterations. Specifically, TMAS maintains two complementary memory banks. The \emph{experience bank} stores low-level, trajectory-specific reasoning signals, including verified intermediate conclusions, concrete problem-solving skills, and verifier-identified errors or pitfalls. It allows later agents to exploit reliable partial progress and avoid repeating local mistakes. The \emph{guideline bank}, in contrast, stores high-level strategic memory distilled from parallel exploration, including global solution directions, key structural insights, and previously explored reasoning strategies. Rather than directly reusing these guidelines, it guides subsequent agents to avoid reproducing previously attempted patterns, thereby promoting non-redundant exploration. Together, these hierarchical memories serve as the communication substrate for multi-agent synergy, enabling specialized agents to share local evidence, propagate global strategies, and convert independent parallel trajectories into a coordinated iterative reasoning process.

\subsection{Multi-Agent Inference System}
As summarized in Algorithm~\ref{alg:TMAS}, TMAS performs inference through an iterative multi-agent exploration pipeline. 
For a given problem $Q$, the system runs for $T$ iterations in total, where each iteration $t$ consists of parallel solution generation, verification, summarization, and memory update. 

\textbf{Solution Generation, Verification, and Summarization.}
At each iteration, a solution agent first generates $N$ candidate solution trajectories in parallel, denoted as $\{c_{t,i}\}_{i=1}^{N}$. 
For each candidate solution $c_{t,i}$, a verification agent performs $M$ independent verification passes, yielding a verification set $\mathcal{V}_{t,i} = \{v_{t,i}^{(m)}\}_{m=1}^{M}$, where each verification $v_{t,i}^{(m)}$ provides both analytical feedback and an associated grading score. 
The resulting verification results are aggregated by a summary agent into a concise rollout-level summary $s_{t,i}$, highlighting validated reasoning steps and potential logical flaws.

\textbf{Memory Update.}
For each candidate at iteration $t$, we define a rollout as $r_{t,i} = (c_{t,i}, s_{t,i})$, and denote the collection of all rollouts as $\mathcal{R}_t = \{r_{t,i}\}_{i=1}^{N}$. 
Given $\mathcal{R}_t$, two memory update agents operate in parallel. The experience agent extracts shared reasoning patterns and reusable intermediate findings across solution trajectories to update the experience bank $\mathcal{E}_t$, while the guideline agent abstracts the high-level solution approaches explored by the parallel rollouts and updates the guideline bank $\mathcal{G}_t$.
The updated experience bank $\mathcal{E}_t$ and guideline bank $\mathcal{G}_t$ are then carried forward to the next iteration, where they serve as part of the conditioning context for subsequent solution generation.

Specifically, TMAS decomposes iterative reasoning into five specialized agents, each responsible for a distinct function in the collaborative inference process. 
We denote them as the solution agent $\mathcal{A}_{\text{sol}}$, verification agent $\mathcal{A}_{\text{ver}}$, summary agent $\mathcal{A}_{\text{sum}}$, experience agent $\mathcal{A}_{\text{exp}}$, and guideline agent $\mathcal{A}_{\text{guide}}$. 
Their roles are defined as follows:
\begin{itemize}[leftmargin=10pt, itemsep=0pt, parsep=3pt, topsep=1pt]
    \item \textit{Solution Agent.} 
    The solution agent $\mathcal{A}_{\text{sol}}$ generates candidate solution trajectories with an exploration coefficient $\epsilon$, where $\epsilon$ controls the balance between exploitation and exploration. 
    At iteration $t$, the $i$-th candidate is sampled as
    \begin{equation}
    c_{t,i} \sim
    \begin{cases}
    \mathcal{A}_{\text{sol}}(Q, \mathcal{R}_{t-1}, \mathcal{E}_{t-1}),
    & \text{with probability } 1-\epsilon,\\
    \mathcal{A}_{\text{sol}}(Q, \mathcal{G}_{t-1}),
    & \text{with probability } \epsilon.
    \end{cases}
    \label{eq:solution-agent}
    \end{equation}
    The first branch exploits previous rollouts and accumulated experience to refine existing reasoning paths, while the second branch encourages non-redundant exploration guided by high-level records of previously explored reasoning routes.

    \item \textit{Verification Agent.} 
    The verification agent $\mathcal{A}_{\text{ver}}$ evaluates each candidate solution $c_{t,i}$ through $M$ independent verification passes, producing a verification set
    \begin{equation}
    \mathcal{V}_{t,i}
    =
    \left\{
    \mathcal{A}_{\text{ver}}^{(m)}(Q, c_{t,i})
    \right\}_{m=1}^{M}.
    \label{eq:verification-agent}
    \end{equation}
    Each verification output provides analytical feedback together with scalar scores that indicate full correctness, partial correctness, or fatal errors.

    \item \textit{Summary Agent.} 
    The summary agent $\mathcal{A}_{\text{sum}}$ aggregates the verification results for each candidate $c_{t,i}$ into a concise summary
    \begin{equation}
    s_{t,i}
    =
    \mathcal{A}_{\text{sum}}(Q, c_{t,i}, \mathcal{V}_{t,i}).
    \label{eq:summary-agent}
    \end{equation}
    This summary consolidates feedback across verification passes, highlighting validated reasoning steps and identifying remaining flaws.

    \item \textit{Experience Agent.} 
    The experience agent $\mathcal{A}_{\text{exp}}$ updates the experience bank $\mathcal{E}_t$ as
    \begin{equation}
    \mathcal{E}_t
    =
    \mathcal{A}_{\text{exp}}(Q, \mathcal{R}_t, \mathcal{E}_{t-1}).
    \label{eq:experience-agent}
    \end{equation}
    It extracts reusable experience from the rollout set $\mathcal{R}_t$, capturing cross-trajectory patterns such as shared intermediate steps and common error-avoidance heuristics.

    \item \textit{Guideline Agent.} 
    The guideline agent $\mathcal{A}_{\text{guide}}$ updates the guideline bank $\mathcal{G}_t$ as
    \begin{equation}
    \mathcal{G}_t
    =
    \mathcal{A}_{\text{guide}}(Q, \mathcal{R}_t, \mathcal{G}_{t-1}).
    \label{eq:guideline-agent}
    \end{equation}
    It abstracts the distinct high-level solution strategies attempted across the parallel rollouts, encouraging more diverse exploration in subsequent iterations.
\end{itemize}
\subsection{Hybrid Reward System with RLVR}

TMAS relies on structured collaboration among multiple agents, where the model must not only generate correct solutions, but also effectively use accumulated memories and continue exploring diverse reasoning paths across iterations. However, standard reinforcement learning with verifiable rewards (RLVR) training mainly optimizes final answer correctness, without explicitly encouraging the model to use accumulated experience or explore beyond previously attempted reasoning routes. To better align the model with the collaborative reasoning process of TMAS, we design a hybrid reward system that jointly preserves basic reasoning capability, enhances experience utilization, and promotes novel strategy exploration.

We implement this training scheme based on GRPO~\cite{deepseekmath}. 
For each training prompt $Q$, GRPO samples $N$ rollouts $\{o_i\}_{i=1}^{N}$ from the old policy $\pi_{\theta_{\mathrm{old}}}$ and optimizes the following clipped objective:
\begin{equation}
J_{\mathrm{GRPO}}(\theta)=
\mathbb{E}_{Q,\{o_i\}}
\left[
\frac{1}{\sum_i |o_i|}
\sum_{i=1}^{N}\sum_{t=1}^{|o_i|}
\min\!\left(
\rho_{i,t}A_i,\,
\operatorname{clip}(\rho_{i,t},1-\epsilon_{\mathrm{low}},1+\epsilon_{\mathrm{high}})A_i
\right)
\right],
\end{equation}
where $\rho_{i,t}=\pi_{\theta}(o_{i,t}\mid Q,o_{i,<t})/\pi_{\theta_{\mathrm{old}}}(o_{i,t}\mid Q,o_{i,<t})$ and $\epsilon_{\mathrm{low}}$ and $\epsilon_{\mathrm{high}}$ are the clipping coefficients. 
The rollout-level advantage is computed by group-normalizing rewards as $A_i=(\tilde r_i-\mu)/(\sigma+\delta)$, where $\mu=\frac{1}{N}\sum_i \tilde r_i$ and $\sigma=\sqrt{\frac{1}{N}\sum_i(\tilde r_i-\mu)^2}$. 
We keep the GRPO objective and advantage normalization unchanged, and only modify the reward $\tilde r_i$ through our hybrid reward system.

Our hybrid reward system consists of three components, corresponding to high-quality solution generation, effective experience utilization, and continued exploration of new reasoning paths.

\textbf{Standard Correctness Reward.}
To preserve the model's core reasoning capability, the first component applies a strict correctness-based reward. 
In this setting, $Q$ corresponds to the standard problem description. 
Each rollout receives $\tilde r_i=1$ if the final answer of $o_i$ is correct, and $\tilde r_i=-1$ otherwise. 
The advantage is then computed using the standard GRPO group normalization.

\textbf{Experience Utilization Reward.}
The goal of this component is to encourage the model to make effective use of the provided experience bank. 
Intuitively, if a problem is difficult to solve using historical trajectories alone but can be solved when the experience bank is provided, then the Bank-conditioned rollout should receive an additional reward. 
This encourages the model to rely on accumulated experience when it provides useful complementary information, rather than treating the experience bank as passive context. We sample $N$ rollouts per prompt and equally partition them into a Base group $\mathcal{B}_{\mathrm{base}}$ and a Bank group $\mathcal{B}_{\mathrm{bank}}$. 
Both groups are conditioned on the same problem $Q$ and historical trajectories, while $\mathcal{B}_{\mathrm{bank}}$ additionally incorporate an experience bank. 
After assigning the standard correctness reward $r_i \in \{+1,-1\}$ on every answer $o_i$, we define the base accuracy as
\begin{equation}
p_{\mathrm{base}}
=
\frac{1}{|\mathcal{B}_{\mathrm{base}}|}
\sum_{i \in \mathcal{B}_{\mathrm{base}}}
\mathbb{I}[r_i=1],
\end{equation}
which serves as a proxy for how well the current problem can be solved without bank information. 
The reward is then reshaped as
\begin{equation}
\tilde r_i=
\begin{cases}
r_i + \beta (1 - p_{\mathrm{base}}), & i \in \mathcal{B}_{\mathrm{bank}},\ r_i=1,\\
r_i, & \text{otherwise},
\end{cases}
\end{equation}
where $\beta$ denotes the maximum bonus coefficient, and $(1-p_{\mathrm{base}})$ modulates this bonus according to the difficulty of solving the problem without the experience bank. 
Thus, correct Bank-group rollouts receive a larger bonus when trajectory-only refinement performs poorly, explicitly encouraging the model to exploit the experience bank in cases where it provides useful additional information. 

\textbf{Novel Strategy Exploration Reward.}
To encourage the discovery of new solution strategies, this component rewards rollouts whose high-level reasoning directions go beyond previously summarized guideline memory. 
For this objective, $Q$ augments the problem description with a set of historical solution guidelines, which provide concise abstractions of previously explored methods rather than full reasoning trajectories. 
For each rollout, we consider two binary signals: correctness and guideline-level novelty. 
Specifically, $r_i \in \{+1,-1\}$ indicates whether the final answer $o_i$ is correct, while $n_i \in \{0,1\}$ indicates whether the solution guideline associated with the rollout is novel. 
Here, $n_i=1$ denotes that the rollout follows a guideline not covered by the existing guideline set, whereas $n_i=0$ denotes that it follows a previously observed guideline.
The reward is defined as
\begin{equation}
\tilde r_i =
\begin{cases}
+1.0, & r_i = +1,\ n_i = 1, \\
+0.2, & r_i = +1,\ n_i = 0, \\
-0.5, & r_i = -1,\ n_i = 1, \\
-1.0, & r_i = -1,\ n_i = 0.
\end{cases}
\end{equation}
This design preserves correctness as the primary objective, while encouraging correct solutions that introduce new high-level guidelines and penalizing repeated or unproductive strategies.

Taken together, these three reward components optimize the model for the TMAS framework by preserving its foundational reasoning ability, enhancing its capacity to leverage accumulated experience, and encouraging exploration beyond previously traversed solution strategies.

\section{Experiments}
\label{sec:exp}
In this section, we present the experimental evaluation of TMAS\footnote{Our code is available at \url{https://github.com/IQuestLab/tmas}.}. We begin by describing the evaluation setup and the training details of our multi-task RL procedure. We then report the main results, followed by ablation studies and further analyses of the system's dynamic behavior.

\subsection{Experimental Setup}

\textbf{Evaluation Setup.}
We primarily evaluate TMAS on two challenging reasoning benchmarks. 
The first is IMO-AnswerBench-50, a filtered subset of IMO-AnswerBench~\cite{imoanswerbench}, with the filtering criteria described in appendix~\ref{sec:appdix-a.2 Evaluation Setup}. 
The second is HLE-Math-100, a mathematics subset of Human's Last Exam~\cite{HLE} extracted and adopted by RSE~\cite{rse}. 
In addition to these main benchmarks, we also conduct evaluations on AIME26 and HMMT-25-Nov. 
Since these two benchmarks are relatively less challenging for the base models considered in our study, we report their results in appendix~\ref{sec:appdix-b.2 Evaluation Results on Additional Benchmarks}. We use Qwen3-30B-A3B-Thinking-2507 and Qwen3-4B-Thinking-2507~\cite{qwen3technicalreport} as the base models. For all experiments, we set the maximum output length to 128K tokens, with \texttt{temperature}=1.0 and \texttt{top\_p}=0.95.
Performance is evaluated using Pass@1 accuracy.

\textbf{Baselines.}
We compare TMAS against several representative test-time scaling methods: (1) \textit{Majority Vote (MV)}~\cite{self-consistency}, a non-iterative baseline that aggregates multiple independently sampled solutions by selecting the most frequent answer. (2) \textit{Self-Refine}~\cite{self-refine}, which iteratively improves solutions based on previous trajectories; (3) \textit{Verify-Refine (V-R)}~\cite{veri-refine}, which uses verification feedback to guide a downstream corrector for iterative refinement; (4) \textit{PaCoRe}~\cite{pacore}, which directly aggregates historical trajectories to support subsequent iterations; and (5) \textit{RSE}~\cite{rse}, which distills raw trajectories into both positive and negative experience signals to improve test-time scaling. The implementation details are provided in appendix~\ref{sec:appdix-a.3 Implementation Details of Baseline Methods}.

\textbf{TMAS configuration.}
For each problem, TMAS runs $N=8$ parallel solution trajectories and uses $M=8$ verification agents to evaluate each trajectory. We set $\epsilon=0.2$ to balance exploration and exploitation, and cap the iteration process at 20 iterations. Detailed prompts for all agents are provided in appendix~\ref{sec:d.1 Prompt Templates for TMAS}. Considering training efficiency, we only use Qwen3-4B-Thinking-2507 as the backbone model for hybrid RL training. TMAS is trained with a batch size of 128 and 16 rollouts per prompt, where each response is allowed up to 80K output tokens. We use a learning rate of $1\times10^{-6}$ and conduct training on 256 NVIDIA H20 GPUs, with FP8 quantization applied to rollout generation. More details of the RL training procedure are provided in the appendix~\ref{sec:appdix-a.4 More RL Training settings}.


\subsection{Data Construction for RL Training}
Our hybrid RL objectives require contextual inference states that are different from standard problem-only RL datasets. Specifically, \textit{Experience Utilization} requires historical trajectories and an experience bank, while \textit{Novel Strategy Exploration} additionally relies on a guideline bank of previously explored reasoning paths. Therefore, before RL training, we construct a cold-start dataset to initialize these task-specific contexts. Starting from open-source RL datasets, including DAPO~\cite{dapo} and Skywork-OR1~\cite{skywork}, we use DeepSeek-V3.2~\cite{deepseekv32} as the teacher model to simulate TMAS-style iterative inference. For each problem, the teacher generates multi-round rollout histories, from which we distill the corresponding experiences and guidelines. This yields training examples that match the contextual input format used by TMAS at test time. Our final training data contains 1.6K instances for \textit{Experience Utilization}, 0.6K for \textit{Novel Strategy Exploration} and 2.2K for \textit{Standard Correctness Reward}.

\subsection{Main Results}
\begin{table}[t]
  \caption{Performance comparison across different methods and representative refinement iterations on IMO-AnswerBench-50 and HLE-Math-100. Performance is measured by Pass@1 accuracy (\%). ``It'' denotes the refinement iteration. Gray entries indicate non-iterative baselines. ``w/ Hybrid-RL'' denotes TMAS using the backbone model further trained with our proposed hybrid reward RL system.}
  \label{tab:main_performance}
  \centering
  \small
  \setlength{\tabcolsep}{4pt}
  \begin{tabular}{l cccccc cccccc}
    \toprule
    \multirow{2}{*}{\textbf{Method}} & \multicolumn{6}{c}{\textbf{IMO-AnswerBench-50}} & \multicolumn{6}{c}{\textbf{HLE-Math-100}} \\
    \cmidrule(r){2-7} \cmidrule(l){8-13}
    & It1 & It3 & It9 & It11 & It17 & It19 & It1 & It3 & It9 & It11 & It17 & It19 \\
    \midrule
    \multicolumn{13}{c}{\textit{\textbf{Qwen3-30B-Thinking-2507}}} \\
    \midrule
    MV@64           & \textcolor{gray}{24.00} & \textcolor{gray}{-} & \textcolor{gray}{-} & \textcolor{gray}{-} & \textcolor{gray}{-} & \textcolor{gray}{-} & \textcolor{gray}{30.30} & \textcolor{gray}{-} & \textcolor{gray}{-} & \textcolor{gray}{-} & \textcolor{gray}{-} & \textcolor{gray}{-} \\
    Self-Refine     & 9.06  & 12.75 & 17.50 & 18.94 & 21.88 & 24.19 & 20.25 & 23.78 & 25.19 & 26.19 & 26.00 & 27.12 \\
    V-R & 10.56 & 16.31 & 24.19 & 25.75 & 30.88 & 31.06 & 18.81 & 21.91 & 29.19 & 29.28 & 29.88 & 30.41 \\
    PaCoRe          & \textbf{26.56} & \textbf{29.31} & 30.31 & 30.25 & 29.44 & 30.31 & \textbf{27.41} & 32.00 & 32.69 & 33.16 & 32.75 & 32.78 \\
    RSE             & 25.31 & 25.38 & 35.12 & 35.31 & 38.38 & 38.00 & 21.28 & 23.31 & 27.47 & 29.97 & 31.13 & 31.75 \\
    TMAS            & 22.06 & 28.56 & \textbf{36.50} & \textbf{37.81} & \textbf{39.31} & \textbf{40.50} & 25.09 & \textbf{33.03} & \textbf{33.22} & \textbf{33.72} & \textbf{35.84} & \textbf{35.38} \\
    \midrule
    \multicolumn{13}{c}{\textit{\textbf{Qwen3-4B-Thinking-2507}}} \\
    \midrule
    MV@64           & \textcolor{gray}{6.00} & \textcolor{gray}{-} & \textcolor{gray}{-} & \textcolor{gray}{-} & \textcolor{gray}{-} & \textcolor{gray}{-} & \textcolor{gray}{15.40} & \textcolor{gray}{-} & \textcolor{gray}{-} & \textcolor{gray}{-} & \textcolor{gray}{-} & \textcolor{gray}{-} \\
    Self-Refine     & 5.50  & 5.44  & 8.31  & 9.25  & 9.31  & 8.88  & 12.12 & 12.19 & 13.47 & 13.53 & 13.31 & 13.39 \\
    V-R & 6.00  & 7.69  & 9.06  & 9.69  & 12.12 & 12.56 & 11.47 & 10.66 & 12.78 & 12.59 & 13.81 & 14.41 \\
    PaCoRe          & 7.62  & 10.75 & 10.94 & 10.19 & 11.12 & 10.94 & 16.09 & 16.25 & 16.25 & 16.53 & 16.38 & 16.47 \\
    RSE             & 11.38 & 13.31 & 11.12 & 12.88 & 15.44 & 16.19 & 16.09 & 16.06 & 13.66 & 14.47 & 12.59 & 15.47 \\
    TMAS            & 6.62  & 12.88 & 15.62 & 14.38 & 17.19 & 17.06 & 15.84 & 16.38 & 16.69 & 17.19 & 17.28 & 17.41 \\
    w/ Hybrid-RL         & \textbf{15.38} & \textbf{22.69} & \textbf{29.25} & \textbf{29.19} & \textbf{30.44} & \textbf{30.88} & \textbf{24.16} & \textbf{25.19} & \textbf{25.09} & \textbf{26.34} & \textbf{27.75} & \textbf{28.16} \\
    \bottomrule
  \end{tabular}
\end{table}

We evaluate TMAS on IMO-AnswerBench-50 and HLE-Math-100 using both Qwen3-30B-A3B-Thinking-2507 and Qwen3-4B-Thinking-2507. To examine how different methods scale with iterative test-time computation, Table~\ref{tab:main_performance} reports representative iterations from the early, intermediate, and late stages, with complete results provided in appendix~\ref{sec:appdix-b.1 Complete Experimental Results}. Based on these results, we draw two key conclusions.

\textbf{TMAS demonstrates stronger iterative scaling ability.} While several baselines achieve competitive performance in the early stage, their improvements tend to slow down or plateau as the number of iterations increases. In contrast, TMAS continues to benefit from additional refinement rounds and achieves the best late-stage performance. For example, with Qwen3-30B-Thinking-2507, TMAS reaches 40.50 on IMO-AnswerBench-50 and 35.38 on HLE-Math-100 at iteration 19, outperforming the strongest iterative baselines at the final stage. These results suggest that multi-agent synergy enables a more effective use of additional test-time computation.

\begin{figure}[t]
    \centering 
    
    \includegraphics[width=\textwidth]{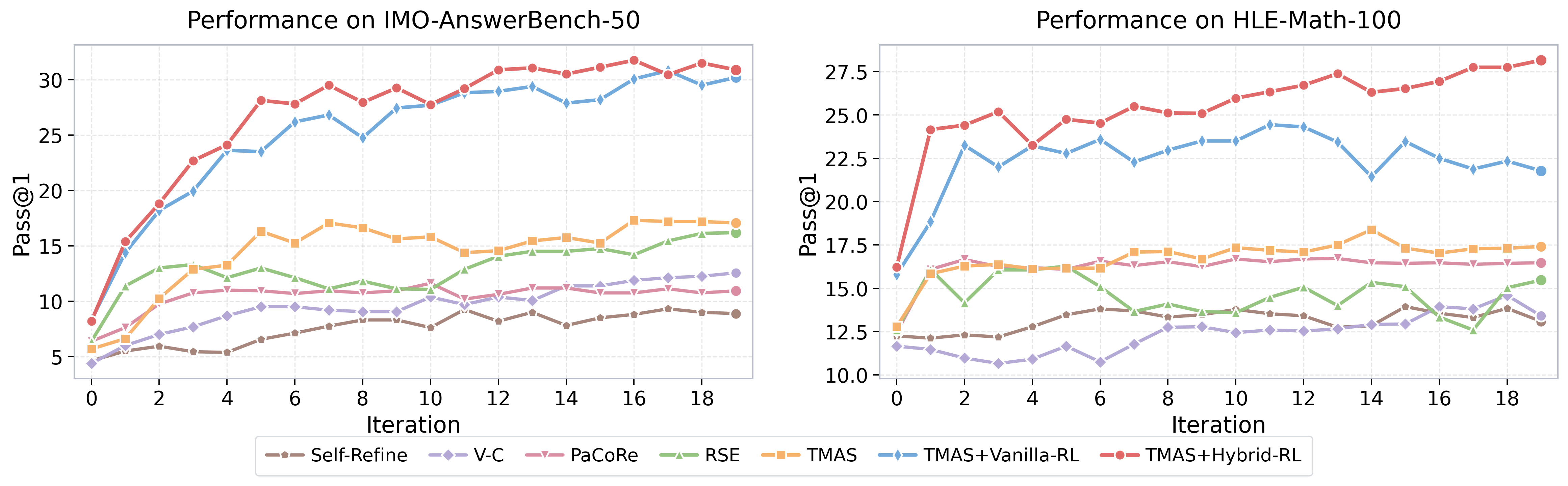}
    
    \caption{Effect of RL training on iterative test-time scaling. TMAS+Vanilla-RL means training solely with standard correctness reward, while TMAS+Hybrid-RL means training with our proposed hybrid reward system. Notably, TMAS+Hybrid-RL not only achieves superior performance from the early stages but also progressively improves as iterations increase, without exhibiting obvious saturation or degradation.} 
    \label{fig:rl-performance} 
\end{figure}

\textbf{Hybrid reward RL unlocks superior and sustained iterative scaling.} Our proposed hybrid reward RL significantly amplifies the model's scaling ability, yielding increasingly pronounced performance advantages as the number of iterations grows. We study this effectiveness using Qwen3-4B-Thinking-2507. As shown in Figure~\ref{fig:rl-performance}, TMAS+Hybrid-RL consistently outperforms TMAS without RL and other iterative baselines across both benchmarks. The improvement is already evident in the first few iterations, indicating that RL provides a stronger initial policy for collaborative iteration. More importantly, TMAS+Hybrid-RL significantly surpasses its counterpart (TMAS+Vanilla-RL) by not only achieving higher peak accuracy but also mitigating the performance degradation observed in Vanilla-RL during later iterations (Figure~\ref{fig:rl-performance} right). This stark contrast demonstrates that our specifically designed experience utilization reward and novel path exploration reward yield distinct advantages in multi-iteration settings, effectively guiding the model to learn a better balance between exploration and exploitation. The model continues to improve with additional iterations, suggesting that RL not only enhances the base reasoning capability but also improves the model's ability to exploit iterative test-time computation. Notably, as a result of this enhancement, at iteration 19, RL substantially narrows the gap between the 4B and 30B TMAS models. The gap is reduced from 23.44 to 9.62 points on IMO-AnswerBench-50 and from 17.97 to 7.22 points on HLE-Math-100, corresponding to relative reductions of 59.0\% and 59.8\%, respectively. These results suggest that, when combined with the TMAS framework, our hybrid RL enables smaller models to approach the performance of substantially larger models through more effective iterative test-time computation.

\subsection{Full-Benchmark Evaluation on IMO-AnswerBench}

To further examine whether the gains observed on the curated hard subset generalize to the complete benchmark, we additionally evaluate TMAS on the full IMO-AnswerBench, which contains 400 olympiad-level problems. 
For consistency with our main experiments, all TMAS-based evaluations on the full benchmark are conducted with a maximum iteration budget of 20. 
The frontier-model results are taken from reported numbers~\cite{kimi-k2.5,deepseekv32} and are included only to provide a reference for the absolute performance of our systems.
As shown in Figure~\ref{fig:full_imo_answerbench}, TMAS substantially improves both Qwen3 backbones on the full benchmark. 
For Qwen3-30B-A3B-Thinking-2507, TMAS improves Pass@1 from 64.9\% to 77.2\%, yielding a 12.3-point gain. 
For Qwen3-4B-Thinking-2507, TMAS combined with Hybrid-RL improves Pass@1 from 53.5\% to 73.5\%, corresponding to a 20.0-point gain. 
These results are consistent with our findings on IMO-AnswerBench-50 and HLE-Math-100: TMAS provides effective multi-agent test-time scaling, while Hybrid-RL further improves the ability of smaller models to exploit accumulated experience and explore new solution strategies.

Notably, the full-benchmark results show that TMAS enables small backbones to reach competitive performance by effectively scaling test-time computation. 
With a 30B backbone, TMAS achieves 77.2\% Pass@1, surpassing the reported result of Opus-4.6 Max and approaching DeepSeek-V3.2, a 685B-scale model. 
More importantly, the 4B model equipped with TMAS and Hybrid-RL reaches 73.5\% Pass@1, substantially narrowing the gap to much larger flagship models. 
Compared with reported large-scale systems such as DeepSeek-V3.2 and Kimi-K2.5, our 4B backbone is much smaller, yet still attains a strong absolute performance level on the full benchmark.

\subsection{Ablation and More Analysis}
In this section, we conduct ablation studies and sensitivity analyses using Qwen3-30B-Thinking-2507 to evaluate the iterative Pass@1 performance of TMAS. We aim to understand not only the complementary contributions of individual modules, but also how delicate trade-offs in exploration coeffcients, verifier feedback, and parallel scaling capacities impact the overall scaling efficiency.

\begin{table*}[htpb]
  \caption{Component ablation study on IMO-AnswerBench-50. ``w/o guideline'', ``w/o experience'', and ``w/o both'' denote TMAS without the guideline module, the experience module, and both modules, respectively. The best result in each iteration is shown in bold.}
  \label{tab:ablation_study}
  \centering
  \small
  \setlength{\tabcolsep}{4pt}
  \begin{tabular}{lcccccccccccc}
    \toprule
    \multirow{2}{*}{\textbf{Method}} & \multicolumn{12}{c}{\textbf{Iteration}} \\
    \cmidrule(l){2-13}
    & It0 & It1 & It2 & It3 & It4 & It5 & It6 & It7 & It8 & It9 & It10 & It11 \\
    \midrule
    TMAS            & \textbf{10.88} & \textbf{22.06} & \textbf{25.94} & \textbf{28.56} & \textbf{31.25} & \textbf{32.56} & \textbf{33.31} & \textbf{33.69} & \textbf{33.88} & \textbf{36.50} & \textbf{35.94} & \textbf{37.81} \\
    w/o guideline   & 6.31           & 10.88          & 14.50          & 24.94          & 28.44          & 28.81          & 30.00          & 29.62          & 31.81          & 33.44          & 34.75          & 35.75          \\
    w/o experience  & 9.44           & 18.38          & 20.94          & 23.06          & 23.06          & 27.56          & 30.88          & 33.38          & 33.00          & 33.75          & 33.50          & 33.44          \\
    w/o both        & 8.61           & 15.05          & 18.69          & 21.36          & 23.02          & 26.08          & 27.87          & 29.53          & 28.89          & 31.51          & 33.16          & 33.16          \\
    \bottomrule
  \end{tabular}
\end{table*}
\textbf{Experience and guidelines drive complementary iterative gains.} To isolate their effects, we perform component ablations on IMO-AnswerBench-50. As shown in Table~\ref{tab:ablation_study}, removing either module degrades Pass@1 performance, with their joint removal (``w/o both'') causing the most severe deterioration. Specifically, the ``w/o guideline'' variant suffers most in early iterations, indicating that guidelines help the model rapidly steer toward promising paths and avoid redundant exploration. In contrast, the ``w/o experience'' variant exhibits weaker gains in later iterations and a lower final accuracy, implying that experience is critical for sustaining effective refinement. These distinct behaviors confirm that the full capability of TMAS relies on the synergy of both modules.

\begin{figure}[t]
    \centering 
    
    \includegraphics[width=\textwidth]{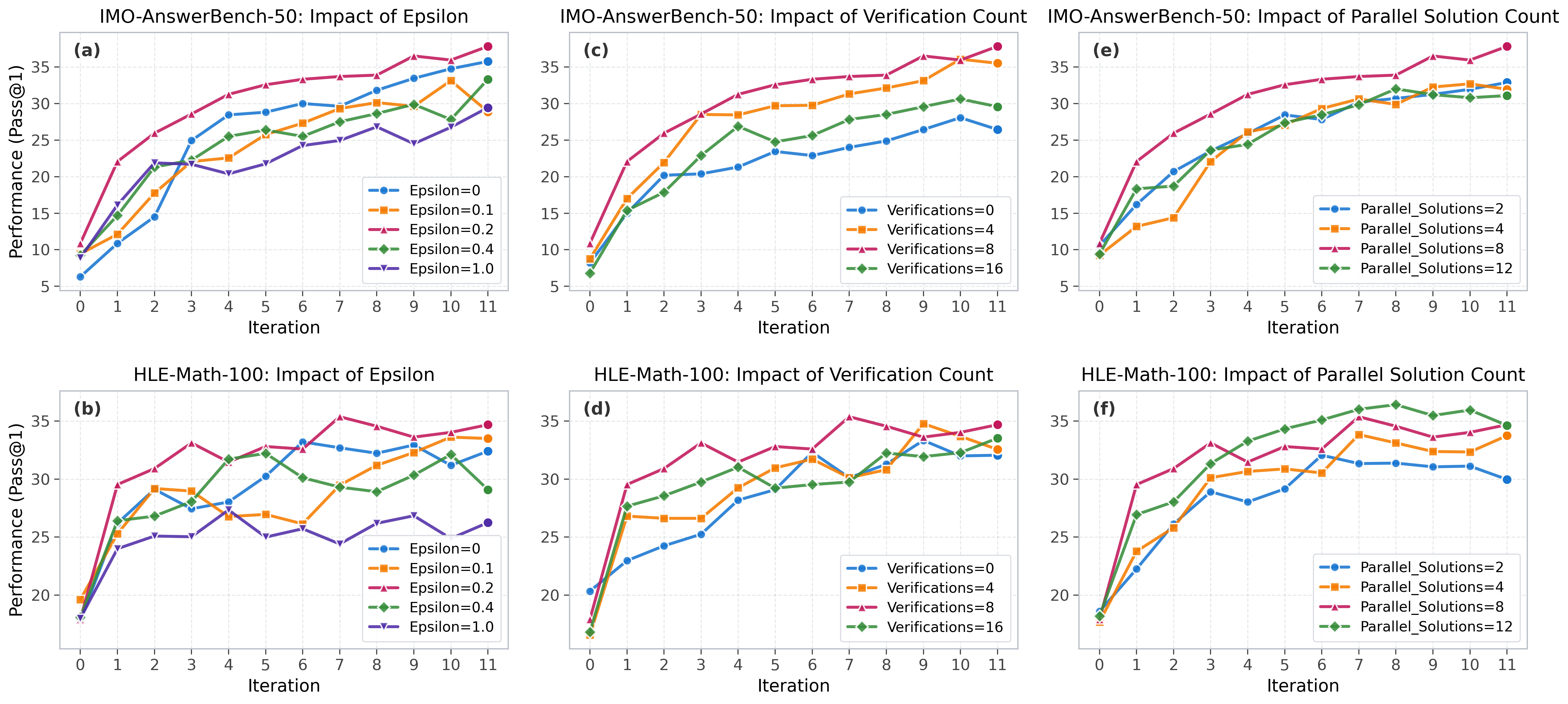}
    
    \caption{Sensitivity analysis of TMAS. Panels (a--b), (c--d), and (e--f) show the impacts of different exploration coefficient $\epsilon$, verification count, and parallel solution count, respectively.} 
    \label{fig:analysis} 
\end{figure}

\textbf{Moderate exploration coefficient optimizes the balance between discovery and exploitation.} We study the effect of the exploration coefficient $\epsilon \in \{0, 0.1, 0.2, 0.4, 1.0\}$ across both benchmarks. As shown in Figure~\ref{fig:analysis}(a) and (b), performance exhibits a non-monotonic trend: both purely exploitative ($\epsilon=0$) and overly exploratory ($\epsilon=1.0$) settings yield suboptimal outcomes. An intermediate value of $\epsilon=0.2$ achieves the highest final Pass@1, confirming that TMAS requires sufficient structural exploration while remaining firmly anchored to successful past trajectories.

\textbf{An optimal verification budget prevents noise and maximizes refinement.} We analyze the effect of varying verification counts per solution among $\{0, 4, 8, 16\}$. As shown in Figure~\ref{fig:analysis}(c) and (d), verification is clearly essential, with its removal ($\text{count}=0$) yielding the lowest Pass@1 accuracy. However, an intermediate count of 8 achieves the best overall results on both benchmarks. Increasing the count to 16 provides no benefit and even degrades performance, suggesting that excessive verification may introduce redundant or inconsistent signals that impair refinement efficiency.

\textbf{Performance gains saturate with excessive parallel solutions.} We examine the impact of the parallel solution budget generated for each problem. As shown in Figure~\ref{fig:analysis}(e) and (f), too few parallel solutions consistently hurt the scaling trajectory by limiting the diversity of reasoning paths. However, increasing the number of solutions does not yield monotonic improvements. While a budget of 8 parallel solutions achieves the highest final Pass@1 on IMO-AnswerBench-50, expanding to 12 solutions brings limited or unstable gains, indicating that additional trajectories beyond this point become difficult for the model to integrate effectively.

\section{Conclusion and Limitations}
\label{sec:conclusion and limitation}

In this work, we propose TMAS, a multi-agent test-time scaling framework that enables structured information flow across agents, trajectories, and iterations. It coordinates solution generation, verification, feedback summarization, experience extraction, and guideline updating into a unified iterative inference process. TMAS introduces hierarchical memory mechanisms to separately maintain low-level experience memory and high-level guideline memory, and further uses a hybrid reward RL scheme to align the model with basic reasoning preservation, experience utilization, and novel strategy exploration. Experiments demonstrate that TMAS achieves stronger iterative scaling than existing TTS baselines, with hybrid reward training further improving scaling effectiveness and stability. However, due to computational and API cost constraints, we have not yet evaluated TMAS on frontier models such as GPT-5.5, where the upper bound of multi-agent test-time synergy could be further examined. In addition, the current RL pipeline requires an external model to pre-construct cold-start trajectories and memory-based training data. Future work can dynamically incorporate trajectories and memory signals from previous iterations into the RL data pool, thereby continuously expanding the data source and better adapting training to the evolving TMAS process.

\clearpage
\newpage

\bibliography{ref}

\clearpage
\newpage

\appendix

\definecolor{promptblue}{RGB}{0,92,170}
\definecolor{promptbg}{RGB}{235,236,250}

\newtcblisting{promptbox}[2][]{%
  enhanced,
  breakable,
  listing only,
  colback=promptbg,
  colframe=promptblue,
  colbacktitle=promptblue,
  coltitle=white,
  fonttitle=\bfseries,
  title={#2},
  boxrule=1pt,
  arc=2mm,
  left=2mm,
  right=2mm,
  top=1mm,
  bottom=1mm,
  listing options={
    basicstyle=\ttfamily\small,
    breaklines=true,
    columns=fullflexible,
    keepspaces=true,
    showstringspaces=false,
    extendedchars=true
  },
  #1
}
\newpage
\section{Experimental Details}
\subsection{Test-Time Inference Algorithm of TMAS}
\label{sec:appdix-a.1 Test-Time Inference Algorithm of TMAS}
The complete test-time inference procedure of TMAS is outlined in Algorithm \ref{alg:TMAS}. During each iteration, the algorithm balances exploration and refinement through an $\epsilon$-greedy generation strategy, and leverages multi-verifier feedback to dynamically update the guidance and exploration states, ultimately returning the candidate with the highest final score.

\begin{algorithm}[ht]
\caption{TMAS test-time inference}
\label{alg:TMAS}
\begin{algorithmic}[1]
\Require problem $x$; iteration budget $T$; number of parallel candidates $N$; number of verifiers $M$; exploration rate $\epsilon$
\State $\mathcal{E}_0 \gets \emptyset$, $\mathcal{G}_0 \gets \emptyset$, $\mathcal{R}_0 \gets \emptyset$

\For{$t = 0$ to $T-1$}
    \State $\mathcal{R}_t \gets \emptyset$
    \Statex \Comment{All $N$ candidates in iteration $t$ are sampled in parallel from the fixed previous state.}
    
    \For{\textbf{parallel} $i = 1$ to $N$}
        \State sample $z_{t,i} \sim \mathrm{Bernoulli}(\epsilon)$
        \If{$z_{t,i}=0$}
            \State $c_{t,i} \gets \mathcal{A}_{sol}\!\left(x,\mathcal{R}_{t-1},\mathcal{E}_{t-1}\right)$
        \Else
            \State $c_{t,i} \gets \mathcal{A}_{sol}\!\left(x,\mathcal{G}_{t-1}\right)$
        \EndIf
        
        \Statex \Comment{Verifier agents can also be executed in parallel.}
        \State $\mathcal{V}_{t,i} \gets \left\{\mathcal{A}_{ver}^{(m)}(x,c_{t,i})\right\}_{m=1}^{M}$ \textbf{in parallel}
        \State $s_{t,i} \gets \mathcal{A}_{sum}\!\left(x,c_{t,i},\mathcal{V}_{t,i}\right)$
        \State $r_{t,i} \gets \left\{\left(c_{t,i}, s_{t,i}\right)\right\}$
    \EndFor

    \State $\mathcal{R}_t \gets \bigcup_{i=1}^{N} r_{t,i}$
    \State $\mathcal{E}_t \gets \mathcal{A}_{exp}\!\left(x,\mathcal{R}_t,\mathcal{E}_{t-1}\right)$
    \State $\mathcal{G}_t \gets \mathcal{A}_{guide}\!\left(x,\mathcal{R}_t,\mathcal{G}_{t-1}\right)$
\EndFor

\State \Return $\arg\max_{c \in \mathcal{R}_T} \mathrm{Score}(c)$
\end{algorithmic}
\end{algorithm}

























\subsection{Evaluation Setup}
\label{sec:appdix-a.2 Evaluation Setup}
The original IMO-AnswerBench contains 400 problems. Evaluating all of them with test-time scaling methods would be computationally expensive and time-consuming. To make the evaluation more efficient while still focusing on sufficiently challenging problems, we construct a smaller evaluation subset using Qwen3-4B-Thinking-2507 as a filter. Specifically, we perform 8 independent inference runs for each problem in the original IMO-AnswerBench and retain the problems that are solved correctly fewer than 2 times out of 8. From this pool of difficult problems, we select 50 problems for evaluation. For HLE-Math-100, we directly adopt the dataset released by RSE~\cite{rse} without modification.

For each generated solution, we use DeepSeek-V3.2 to assess its correctness against the reference answer in 4 independent judgment runs. This produces 4 binary correctness labels for each solution. We then use these judgment results to compute Pass@1 accuracy by averaging the proportion of solutions judged correct over all problems, and finally averaging across the 4 judgment runs.

The Pass@1 accuracy used in our evaluation is calculated using the following formula:
\[
\text{Pass@1}^{(r)} = \frac{1}{|\mathcal{P}|} \sum_{i=1}^{|\mathcal{P}|} \frac{c_i^{(r)}}{n_i},
\]
where \(|\mathcal{P}|\) denotes the total number of problems, \(n_i\) is the number of sampled rollouts for problem \(i\), and \(c_i^{(r)}\) is the number of rollouts judged correct for problem \(i\) under the \(r\)-th judge run. When multiple judge runs are used, we report the average over all judge runs:
\[
\text{Pass@1} = \frac{1}{R} \sum_{r=1}^{R} \text{Pass@1}^{(r)},
\]
where \(R\) is the total number of judge runs.

The evaluation prompt we use for LLM-as-Judge is listed below:

\begin{promptbox}{LLM-as-Judge System Prompt}
You are an expert Math Evaluator. Your task is to verify if the [Student's Answer] is mathematically equivalent to the [Reference Answer] for the given [Problem].
Rules for Equivalence:
1. Numerical: 0.5 is equivalent to 1/2. 1000 is equivalent to 1,000.
2. Algebraic: x+1 is equivalent to 1+x. \frac{1}{\sqrt{2}} is equivalent to \frac{\sqrt{2}}{2}.
3. Formatting: Ignore Markdown formatting (bold, italic), latex styling (\text{}, \mathrm{}), or whitespace differences.
4. Content: Focus ONLY on the final result value. Ignore the student's reasoning steps unless the result is embedded within them.
5. Units: If the reference implies units and the student omits them (or vice versa) but the number is correct, count it as correct unless the problem explicitly demands unit conversion.

Output Format:
Respond strictly in JSON format. Do not output markdown code blocks.
\end{promptbox}

\begin{promptbox}{LLM-as-Judge User Prompt}
<problem>
{problem}
</problem>

<reference>
{reference}
</reference>

<student_answer>
{student_answer}
</student_answer>

[Task]
Compare the Student's Answer to the Reference Answer. 
1. Analyze the mathematical value of both answers.
2. Determine if they represent the same solution (equivalent).
3. If the student answer contains a derivation, look for the final result.

Respond in JSON:
{
    "reasoning": "Brief explanation...",
    "equivalent": true/false
}
\end{promptbox}

\subsection{Implementation Details of Baseline Methods}
\label{sec:appdix-a.3 Implementation Details of Baseline Methods}
For Self-Refine, we generate 8 solutions in parallel, and each solution is refined independently in subsequent rounds without any interaction across different trajectories. For the Verify-Refine (V-R) baseline, we first generate a solution and then apply a verifier to assess it. After that, a corrector is used to generate a revised solution based on both the preceding solution and the corresponding verification result. For PaCoRe and RSE, we directly use their official implementations, with only minor modifications to the input format to accommodate different datasets. For a fair comparison, we set \texttt{num\_responses\_per\_round}=8 and \texttt{N\_COMPLETIONS}=8 for PaCoRe, matching TMAS's parallel solution budget of 8.

\subsection{More RL Training settings}
\label{sec:appdix-a.4 More RL Training settings}
We conduct hybrid RL training using Qwen3-4B-Thinking-2507 as the backbone model. Training is implemented with verl~\cite{verl}. During training, each prompt is sampled with 16 rollouts, and each response is allowed up to 80K output tokens. We train the model on 256 NVIDIA H20 GPUs. The training batch size is set to 128. We use a learning rate of $1\times10^{-6}$ and train for 190 steps. To support long-output RL training, we enable dynamic batching, optimizer offloading, and activation recomputation. Rollout generation is performed with SGLang, and FP8 quantization is applied during rollout generation to improve efficiency. We set clipping range to $[0.20, 0.255]$, with the clipping coefficient set to 10.0. We set the entropy coefficient to 0 and do not apply KL regularization in the reward. For our proposed \textit{Experience Utilization Reward}, we set the maximum bonus coefficient $\beta=0.6$.

\section{More Experiment Results and Analysis}
\subsection{Complete Experimental Results}
\label{sec:appdix-b.1 Complete Experimental Results}
In this section, we provide the comprehensive, iteration-by-iteration numerical results that supplement the main experimental findings. Tables~\ref{tab:app_qwen30b_imo} to \ref{tab:app_qwen4b_hle_vanilla_rl} detail the precise performance trajectories of all evaluated methods across the full inference budget of 20 iterations (from It0 to It19). 

Specifically, the results are organized as follows:
\begin{itemize}[leftmargin=10pt]
    \item Tables~\ref{tab:app_qwen30b_imo} and \ref{tab:app_qwen30b_hle}report the detailed baseline comparisons using the larger Qwen3-30B-A3B-Thinking-2507 model on the IMO-AnswerBench-50 and HLE-Math-100 datasets, respectively.
    \item Tables~\ref{tab:app_qwen4b_imo_base} and \ref{tab:app_qwen4b_hle_base} present the corresponding baseline comparisons for the smaller Qwen3-4B-Thinking-2507-Thinking model.
    \item Tables~\ref{tab:app_qwen4b_imo_rl} and \ref{tab:app_qwen4b_hle_rl} break down the impact of our hybrid reward RL training approach on the Qwen3-4B-Thinking-2507 model, tracking the performance variations across different training checkpoints (from No RL up to Step-190).
    \item Tables~\ref{tab:app_qwen4b_imo_vanilla_rl} and \ref{tab:app_qwen4b_hle_vanilla_rl} further report the performance of Vanilla-RL on the Qwen3-4B-Thinking-2507 model after 190 training steps, where Vanilla-RL denotes RL training using only the correctness reward without our proposed hybrid reward design.
\end{itemize}

These detailed breakdowns are intended to give a transparent view of the step-by-step scaling behavior of test-time compute. As shown across the tables, while baseline methods often plateau or degrade after a certain number of steps, our proposed pipeline consistently maintains positive scaling and achieves superior peak performance as the iteration count increases.

\begingroup 
\setlength{\intextsep}{2pt} 
\begin{table}[H]
\vspace{4pt}
  \caption{Detailed performance of Qwen3-30B-A3B-Thinking-2507 baseline methods on IMO-AnswerBench-50 across all iterations (0-19).}
  \label{tab:app_qwen30b_imo}
  \centering
  \resizebox{\textwidth}{!}{
  \begin{tabular}{l cccccccccccccccccccc}
    \toprule
    \textbf{Method} & It0 & It1 & It2 & It3 & It4 & It5 & It6 & It7 & It8 & It9 & It10 & It11 & It12 & It13 & It14 & It15 & It16 & It17 & It18 & It19 \\
    \midrule
    Self-Refine     & 8.31  & 9.06  & 10.31 & 12.75 & 13.88 & 14.75 & 16.06 & 16.50 & 18.12 & 17.50 & 18.19 & 18.94 & 19.12 & 20.25 & 21.56 & 21.75 & 21.62 & 21.88 & 23.75 & 24.19 \\
    V-R & 8.19  & 10.56 & 13.94 & 16.31 & 17.94 & 19.38 & 21.00 & 22.38 & 23.06 & 24.19 & 24.38 & 25.75 & 28.13 & 29.25 & 28.94 & 29.50 & 30.88 & 30.88 & 31.31 & 31.06 \\
    PaCoRe          & 11.56 & 26.56 & 26.50 & 29.31 & 29.62 & 29.69 & 29.19 & 30.50 & 30.12 & 30.31 & 29.94 & 30.25 & 30.31 & 30.69 & 30.62 & 29.25 & 30.06 & 29.44 & 30.31 & 30.31 \\
    RSE             & 10.50 & 25.31 & 24.06 & 25.38 & 27.81 & 33.00 & 34.50 & 31.88 & 35.00 & 35.12 & 36.19 & 35.31 & 37.25 & 38.12 & 36.38 & 37.31 & 37.25 & 38.38 & 38.12 & 38.00 \\
    TMAS   & 10.88 & 22.06 & 25.94 & 28.56 & 31.25 & 32.56 & 33.31 & 33.69 & 33.88 & 36.50 & 35.94 & 37.81 & 35.81 & 35.38 & 38.06 & 36.81 & 37.81 & 39.31 & 41.44 & 40.50 \\
    \bottomrule
  \end{tabular}
  }
\end{table}

\begin{table}[H]
\vspace{8pt}
  \caption{Detailed performance of Qwen3-30B-A3B-Thinking-2507 baseline methods on HLE-Math-100 across all iterations (0-19).}
  \label{tab:app_qwen30b_hle}
  \centering
  \resizebox{\textwidth}{!}{
  \begin{tabular}{l cccccccccccccccccccc}
    \toprule
    \textbf{Method} & It0 & It1 & It2 & It3 & It4 & It5 & It6 & It7 & It8 & It9 & It10 & It11 & It12 & It13 & It14 & It15 & It16 & It17 & It18 & It19 \\
    \midrule
    Self-Refine     & 19.31 & 20.25 & 22.72 & 23.78 & 24.59 & 23.84 & 24.66 & 25.38 & 25.53 & 25.19 & 26.22 & 26.19 & 26.75 & 26.28 & 26.59 & 26.84 & 26.88 & 26.00 & 26.72 & 27.12 \\
    V-R & 16.84 & 18.81 & 20.81 & 21.91 & 23.78 & 25.09 & 26.53 & 26.75 & 27.12 & 29.19 & 28.22 & 29.28 & 30.00 & 29.09 & 29.25 & 29.97 & 30.78 & 29.88 & 30.06 & 30.41 \\
    PaCoRe          & 22.75 & 27.41 & 29.66 & 32.00 & 32.00 & 32.25 & 32.28 & 32.59 & 32.78 & 32.69 & 33.31 & 33.16 & 33.28 & 32.84 & 33.19 & 33.00 & 32.91 & 32.75 & 32.59 & 32.78 \\
    RSE             & 21.06 & 21.28 & 25.38 & 23.31 & 24.44 & 24.00 & 26.88 & 27.44 & 27.78 & 27.47 & 28.66 & 29.97 & 30.59 & 30.75 & 30.56 & 30.41 & 30.66 & 31.13 & 31.72 & 31.75 \\
    TMAS   & 20.72 & 25.09 & 27.00 & 33.03 & 33.40 & 33.52 & 33.74 & 33.30 & 35.97 & 33.22 & 35.05 & 33.72 & 34.41 & 34.12 & 35.81 & 35.81 & 34.91 & 35.84 & 36.81 & 35.38 \\
    \bottomrule
  \end{tabular}
  }
\end{table}

\begin{table}[H]
\vspace{8pt}
  \caption{Detailed performance of Qwen3-4B-Thinking-2507 baseline methods on IMO-AnswerBench-50 across all iterations (0-19).}
  \label{tab:app_qwen4b_imo_base}
  \centering
  \resizebox{\textwidth}{!}{
  \begin{tabular}{l cccccccccccccccccccc}
    \toprule
    \textbf{Method} & It0 & It1 & It2 & It3 & It4 & It5 & It6 & It7 & It8 & It9 & It10 & It11 & It12 & It13 & It14 & It15 & It16 & It17 & It18 & It19 \\
    \midrule
    Self-Refine     & 4.62  & 5.50  & 5.94  & 5.44  & 5.38  & 6.56  & 7.12  & 7.75  & 8.31  & 8.31  & 7.62  & 9.25  & 8.19  & 9.00  & 7.81  & 8.50  & 8.81  & 9.31  & 9.00  & 8.88  \\
    V-R & 4.38  & 6.00  & 7.00  & 7.69  & 8.69  & 9.50  & 9.50  & 9.19  & 9.06  & 9.06  & 10.38 & 9.69  & 10.38 & 10.06 & 11.38 & 11.38 & 11.88 & 12.12 & 12.25 & 12.56 \\
    PaCoRe          & 6.38  & 7.62  & 9.75  & 10.75 & 11.00 & 10.94 & 10.69 & 10.94 & 10.75 & 10.94 & 11.62 & 10.19 & 10.62 & 11.19 & 11.19 & 10.75 & 10.75 & 11.12 & 10.75 & 10.94 \\
    RSE             & 6.38  & 11.38 & 13.00 & 13.31 & 12.12 & 13.00 & 12.12 & 11.12 & 11.81 & 11.12 & 11.06 & 12.88 & 14.06 & 14.50 & 14.50 & 14.75 & 14.19 & 15.44 & 16.12 & 16.19 \\
    TMAS   & 5.69  & 6.62  & 10.25 & 12.88 & 13.25 & 16.31 & 15.25 & 17.06 & 16.62 & 15.62 & 15.81 & 14.38 & 14.56 & 15.44 & 15.75 & 15.25 & 17.31 & 17.19 & 17.19 & 17.06 \\
    \bottomrule
  \end{tabular}
  }
\end{table}

\begin{table}[H]
\vspace{8pt}
  \caption{Detailed performance of Qwen3-4B-Thinking-2507 baseline methods on HLE-Math-100 across all iterations (0-19).}
  \label{tab:app_qwen4b_hle_base}
  \centering
  \resizebox{\textwidth}{!}{
  \begin{tabular}{l cccccccccccccccccccc}
    \toprule
    \textbf{Method} & It0 & It1 & It2 & It3 & It4 & It5 & It6 & It7 & It8 & It9 & It10 & It11 & It12 & It13 & It14 & It15 & It16 & It17 & It18 & It19 \\
    \midrule
    Self-Refine     & 12.25 & 12.12 & 12.31 & 12.19 & 12.78 & 13.47 & 13.81 & 13.69 & 13.34 & 13.47 & 13.78 & 13.53 & 13.41 & 12.75 & 12.84 & 13.94 & 13.56 & 13.31 & 13.84 & 13.39 \\
    V-R & 11.66 & 11.47 & 10.97 & 10.66 & 10.91 & 11.66 & 10.75 & 11.78 & 12.75 & 12.78 & 12.44 & 12.59 & 12.53 & 12.66 & 12.91 & 12.94 & 13.94 & 13.81 & 14.59 & 14.41 \\
    PaCoRe          & 12.44 & 16.09 & 16.66 & 16.25 & 16.19 & 16.09 & 16.56 & 16.31 & 16.53 & 16.25 & 16.69 & 16.53 & 16.69 & 16.72 & 16.47 & 16.44 & 16.47 & 16.38 & 16.44 & 16.47 \\
    RSE             & 12.59 & 16.09 & 14.16 & 16.06 & 16.06 & 16.28 & 15.09 & 13.66 & 14.09 & 13.66 & 13.59 & 14.47 & 15.06 & 14.00 & 15.34 & 15.09 & 13.34 & 12.59 & 15.03 & 15.47 \\
    TMAS   & 12.78 & 15.84 & 16.28 & 16.38 & 16.12 & 16.16 & 16.16 & 17.09 & 17.12 & 16.69 & 17.34 & 17.19 & 17.09 & 17.50 & 18.38 & 17.31 & 17.03 & 17.28 & 17.31 & 17.41 \\
    \bottomrule
  \end{tabular}
  }
\end{table}

\begin{table}[H]
\vspace{8pt}
  \caption{Performance of our RL training approach on Qwen3-4B-Thinking-2507 on IMO-AnswerBench-50 across training steps.}
  \label{tab:app_qwen4b_imo_rl}
  \centering
  \resizebox{\textwidth}{!}{
  \begin{tabular}{l cccccccccccccccccccc}
    \toprule
    \textbf{Method} & It0 & It1 & It2 & It3 & It4 & It5 & It6 & It7 & It8 & It9 & It10 & It11 & It12 & It13 & It14 & It15 & It16 & It17 & It18 & It19 \\
    \midrule
    TMAS (No RL)    & 5.69  & 6.62  & 10.25 & 12.88 & 13.25 & 16.31 & 15.25 & 17.06 & 16.62 & 15.62 & 15.81 & 14.38 & 14.56 & 15.44 & 15.75 & 15.25 & 17.31 & 17.19 & 17.19 & 17.06 \\
    TMAS (Step-100) & 6.81  & 13.19 & 12.12 & 14.44 & 15.88 & 20.44 & 23.44 & 24.25 & 23.25 & 23.94 & 25.38 & 26.31 & 24.88 & 26.62 & 25.31 & 26.25 & 25.88 & 25.50 & 26.38 & 26.25 \\
    TMAS (Step-140) & 8.44  & 17.25 & 16.88 & 16.00 & 16.81 & 16.69 & 20.44 & 22.88 & 25.88 & 25.81 & 29.12 & 30.00 & 29.44 & 29.56 & 30.12 & 32.44 & 33.12 & 32.00 & 30.75 & 30.50 \\
    TMAS (Step-190) & 8.19  & 15.38 & 18.81 & 22.69 & 24.12 & 28.12 & 27.81 & 29.50 & 27.94 & 29.25 & 27.75 & 29.19 & 30.88 & 31.06 & 30.50 & 31.12 & 31.75 & 30.44 & 31.50 & 30.88 \\
    \bottomrule
  \end{tabular}
  }
\end{table}

\begin{table}[H]
\vspace{8pt}
  \caption{Performance of our RL training approach on Qwen3-4B-Thinking-2507 on HLE-Math-100 across training steps.}
  \label{tab:app_qwen4b_hle_rl}
  \centering
  \resizebox{\textwidth}{!}{
  \begin{tabular}{l cccccccccccccccccccc}
    \toprule
    \textbf{Method} & It0 & It1 & It2 & It3 & It4 & It5 & It6 & It7 & It8 & It9 & It10 & It11 & It12 & It13 & It14 & It15 & It16 & It17 & It18 & It19 \\
    \midrule
    TMAS (No RL)    & 12.78 & 15.84 & 16.28 & 16.38 & 16.12 & 16.16 & 16.16 & 17.09 & 17.12 & 16.69 & 17.34 & 17.19 & 17.09 & 17.50 & 18.38 & 17.31 & 17.03 & 17.28 & 17.31 & 17.41 \\
    TMAS (Step-100) & 11.47 & 12.62 & 15.53 & 14.47 & 13.75 & 14.41 & 14.00 & 14.34 & 13.47 & 12.75 & 13.22 & 15.50 & 14.81     & 15.00     & 14.88     & 16.12     & 17.06     & 17.41     & 16.72     & 16.88     \\
    TMAS (Step-140) & 14.47 & 17.88 & 17.28 & 19.31 & 18.84 & 19.75 & 20.28 & 19.53 & 19.91 & 19.72 & 20.62 & 20.88 & 20.28 & 20.75 & 20.44 & 20.16 & 19.88 & 19.97 & 19.31 & 19.56 \\
    TMAS (Step-190) & 16.22 & 24.16 & 24.41 & 25.19 & 23.25 & 24.75 & 24.53 & 25.50 & 25.12 & 25.09 & 25.97 & 26.34 & 26.72 & 27.38 & 26.31 & 26.53 & 26.94 & 27.75 & 27.75 & 28.16 \\
    \bottomrule
  \end{tabular}
  }
\end{table}

\begin{table}[H]
\vspace{8pt}
  \caption{Performance of Vanilla-RL on Qwen3-4B-Thinking-2507 on IMO-AnswerBench-50 across iterations.}
  \label{tab:app_qwen4b_imo_vanilla_rl}
  \centering
  \resizebox{\textwidth}{!}{
  \begin{tabular}{l cccccccccccccccccccc}
    \toprule
    \textbf{Method} & It0 & It1 & It2 & It3 & It4 & It5 & It6 & It7 & It8 & It9 & It10 & It11 & It12 & It13 & It14 & It15 & It16 & It17 & It18 & It19 \\
    \midrule
    TMAS (Vanilla-RL) & 8.31 & 14.38 & 18.19 & 19.94 & 23.62 & 23.50 & 26.19 & 26.81 & 24.75 & 27.44 & 27.69 & 28.81 & 28.94 & 29.38 & 27.88 & 28.19 & 30.06 & 30.81 & 29.50 & 30.19 \\
    \bottomrule
  \end{tabular}
  }
\end{table}

\begin{table}[H]
\vspace{8pt}
  \caption{Performance of Vanilla-RL on Qwen3-4B-Thinking-2507 on HLE-Math-100 across iterations.}
  \label{tab:app_qwen4b_hle_vanilla_rl}
  \centering
  \resizebox{\textwidth}{!}{
  \begin{tabular}{l cccccccccccccccccccc}
    \toprule
    \textbf{Method} & It0 & It1 & It2 & It3 & It4 & It5 & It6 & It7 & It8 & It9 & It10 & It11 & It12 & It13 & It14 & It15 & It16 & It17 & It18 & It19 \\
    \midrule
    TMAS (Vanilla-RL) & 15.78 & 18.84 & 23.25 & 22.00 & 23.22 & 22.78 & 23.59 & 22.28 & 22.97 & 23.50 & 23.50 & 24.44 & 24.31 & 23.44 & 21.44 & 23.47 & 22.50 & 21.88 & 22.34 & 21.78 \\
    \bottomrule
  \end{tabular}
  }
\end{table}
\endgroup 
\subsection{Evaluation Results on Additional Benchmarks}
\label{sec:appdix-b.2 Evaluation Results on Additional Benchmarks}
In addition to IMO-AnswerBench50 and HLE-Math-100, we also evaluate our TMAS pipeline on AIME26 and HMMT-25-Nov. We use Qwen3-30B-A3B-Thinking-2507 as the base model and run each method for 12 iterations.
\begin{figure}[htbp]
    \centering
    \includegraphics[width=\textwidth]{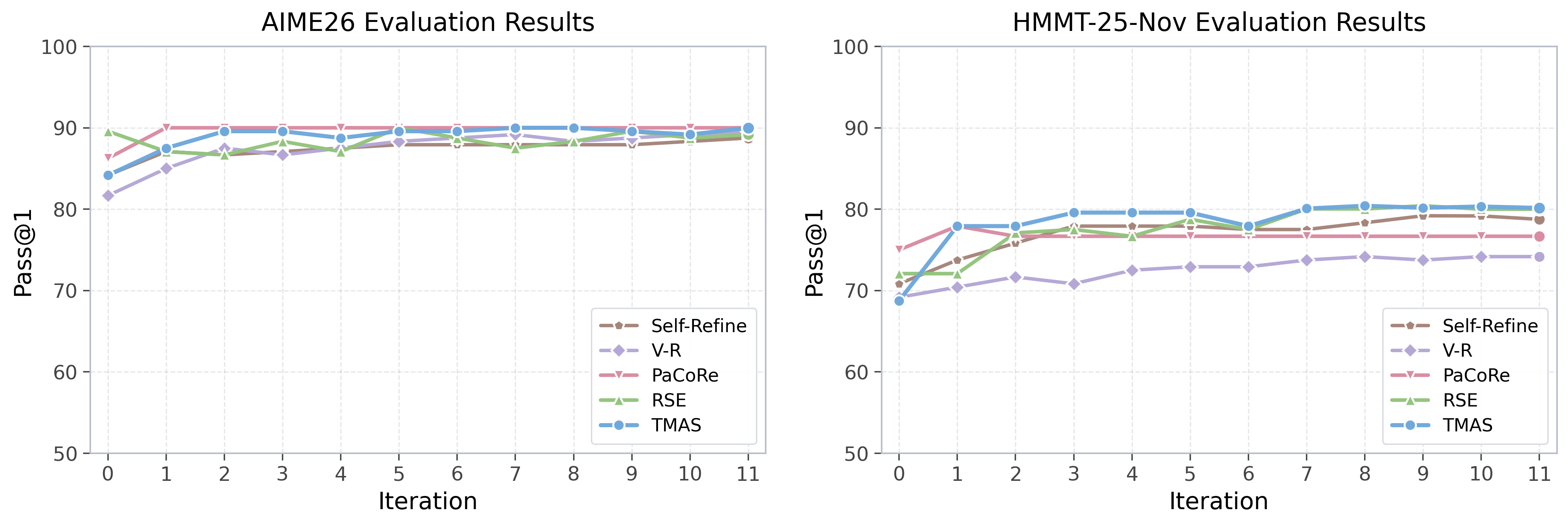}
    \caption{Evaluation results on AIME26 and HMMT-25-Nov over 12 iterations.}
    \label{fig:30b-aime-hmmt}
\end{figure}
We treat these two benchmarks as supplementary evaluations rather than main results. Our reason is that AIME26 and HMMT-25-Nov appear to be relatively easy for our base model, which often already achieve high scores on them. As a result, these benchmarks are less aligned with the setting that test-time computing is primarily designed for, namely, improving performance on genuinely challenging problems. Consistent with this intuition, the results in Figure~\ref{fig:30b-aime-hmmt} show that the performance differences among different methods are very small on both benchmarks.

\subsection{The Impact of Different Exploration Levels}
\label{sec:appdix-b.3 The Impact of Different Exploration Levels}
\begin{figure}[htbp]
    \centering 
    
    \includegraphics[width=0.6\textwidth]{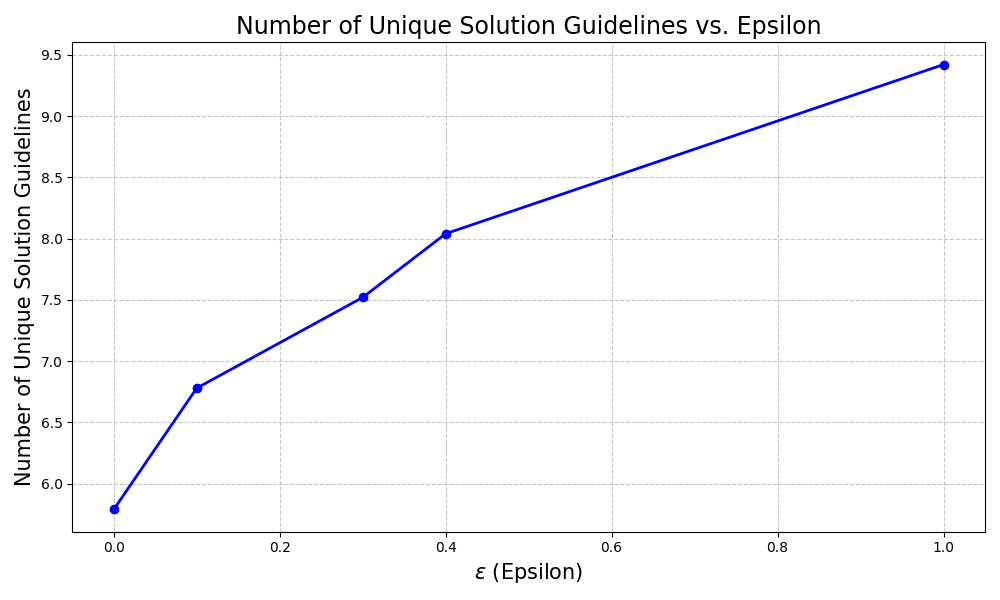}
    
    \caption{Relationship between the exploration coefficient and the total count of unique solution guidelines.} 
    \label{fig:eps-guide} 
\end{figure}
We further study the effect of the exploration factor $\epsilon$ on the diversity of test-time scaling. Specifically, we set exploration  coefficient $\epsilon \in \{0, 0.1, 0.3, 0.4, 1.0\}$ and measure the average number of unique solution guidelines per problem on IMO-AnswerBench-50. As illustrated in Figure~\ref{fig:eps-guide}, larger values of $\epsilon$ lead to a larger number of unique solution guidelines. This result suggests that increasing $\epsilon$ encourages the model to explore more diverse reasoning paths during inference, which is consistent with the intended role of the exploration mechanism in TMAS.

\subsection{The Paradox of Verification: A Shared Capability Boundary}
\label{sec:appdix-b.4 The Paradox of Verification: A Shared Capability Boundary}

To understand the bottlenecks of Test-Time Scaling (TTS) on the hardest problems, we investigate the reliability of the verification signal during iterative refinement in TMAS. 
We partition the IMO-AnswerBench-50 problems into two groups: \emph{ever-correct}, where at least one correct solution is found across all iterations and rollouts, and \emph{never-correct}, where no correct solution is found. 
We compare the base Qwen3-4B-Thinking-2507 model against its counterpart enhanced by our proposed TMAS-oriented RL training.

\begin{figure}[htbp]
    \centering
    \includegraphics[width=0.9\linewidth]{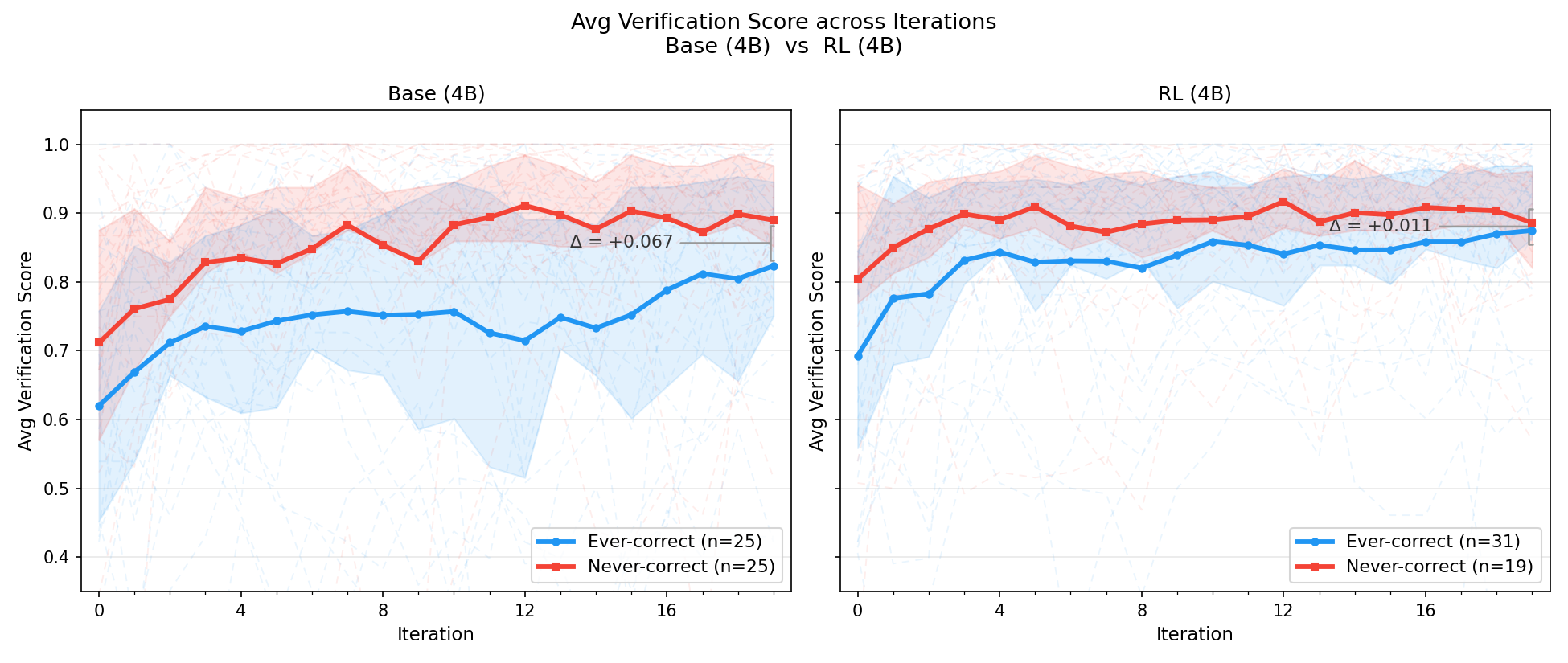}
    \caption{
    Verification score dynamics across TTS iterations on IMO-AnswerBench-50. 
    Problems are stratified into ever-correct and never-correct groups according to whether at least one correct solution is found across all iterations and rollouts. 
    Left: base model. Right: RL-enhanced model under the TMAS framework. 
    Faint dashed lines denote individual problem trajectories, shaded regions denote the interquartile range across problems within each group, and solid lines denote group means. 
    In the base model, never-correct problems receive persistently higher verification scores than ever-correct problems, indicating a shared capability boundary between solution generation and verification. 
    After TMAS-oriented RL training, both groups shift toward higher verification scores and become closer, but the verification signal remains weakly discriminative near the new capability frontier.
    }
    \label{fig:app-verification-iters}
\end{figure}

\begin{figure}[htbp]
    \centering
    \includegraphics[width=0.85\linewidth]{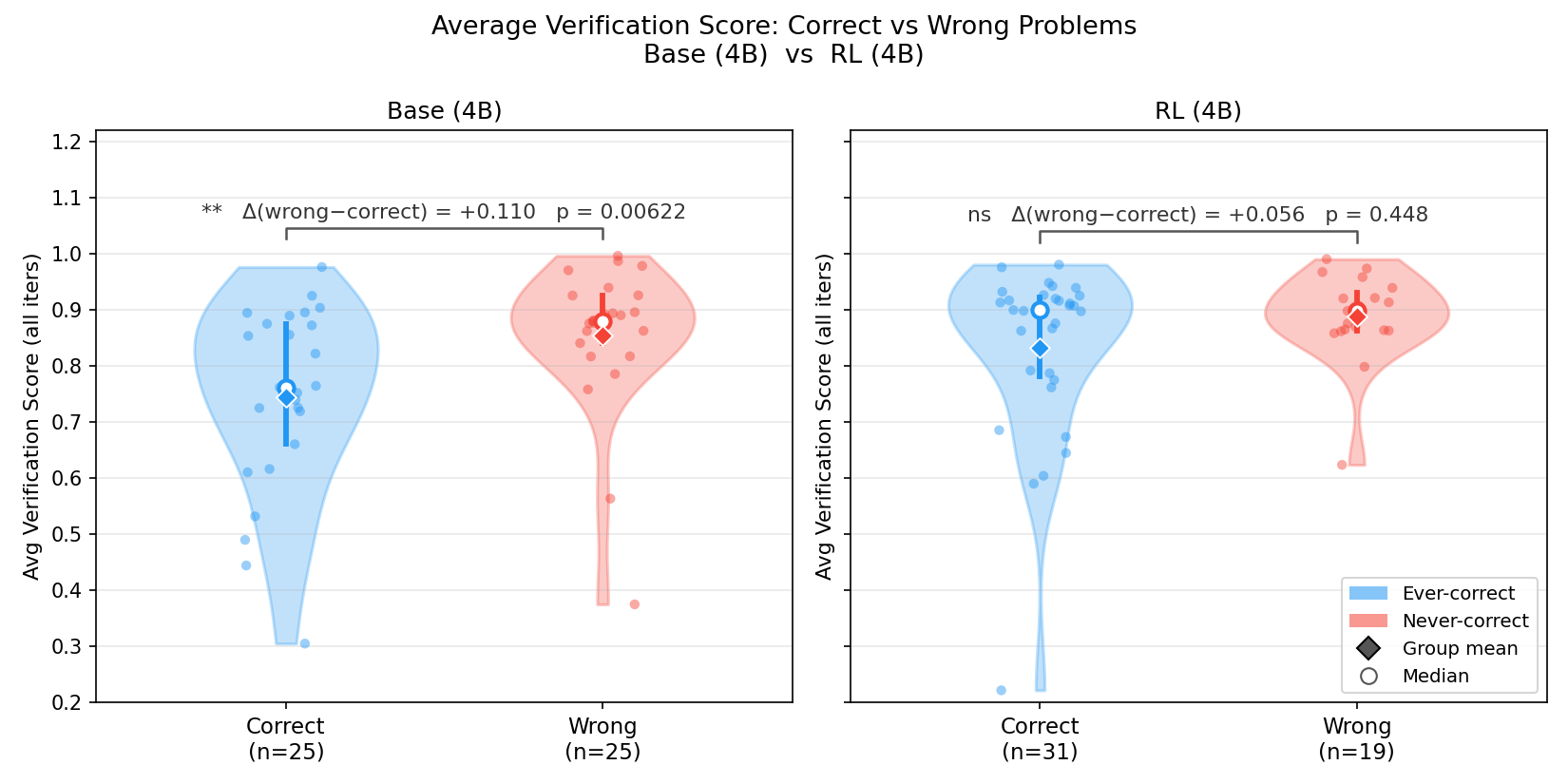}
    \caption{
    Distribution of per-problem average verification scores on IMO-AnswerBench-50. 
    Each point represents one problem, with the score averaged over all TTS iterations and rollouts. 
    ``Correct'' denotes the ever-correct group, and ``Wrong'' denotes the never-correct group. 
    Diamonds indicate group means, hollow circles indicate medians, and vertical bars indicate interquartile ranges. 
    For the base model, never-correct problems receive significantly higher verification scores than ever-correct problems 
    ($\Delta=+0.110$, $p=0.00622$), showing that the hardest unsolved problems are also difficult for the verification agent to assess reliably. 
    After TMAS-oriented RL training, the gap becomes smaller and statistically non-significant 
    ($\Delta=+0.056$, $p=0.448$), while both groups receive generally higher scores.
    }
    \label{fig:app-verification-violin}
\end{figure}

Our analysis reveals a counter-intuitive paradox in the base model: the verification agent assigns higher scores to problems that the solution agent cannot solve. 
As shown in Figure~\ref{fig:app-verification-iters} (Left), the never-correct group consistently receives higher average verification scores than the ever-correct group across all TTS iterations. 
Figure~\ref{fig:app-verification-violin} (left) further quantifies this phenomenon across all rollouts. 
For the base model, the mean verification score for never-correct problems is $0.854$, which is significantly higher than that for ever-correct problems ($0.744$), yielding 
$\Delta(\mathrm{wrong}-\mathrm{correct}) = +0.110$ 
(Mann--Whitney U test, $p = 0.00622$).


Rather than a localized failure of the verification agent, this pattern reflects a shared capability boundary between solution generation and verification. 
On problems exceeding the model's reasoning capacity, the solution agent may still produce plausible and well-structured but ultimately flawed solutions. 
At the same time, because the verification agent is instantiated from the same underlying model family and operates near the same reasoning frontier, it may also lack the capability to identify the specific flaw that invalidates the solution. 
Consequently, incorrect solutions to hard problems can receive high verification scores, depriving the TMAS refinement loop of the reliable discriminative feedback needed for effective correction. After applying our TMAS-oriented RL training, the overall distribution of verification scores shifts upward, as shown in Figure~\ref{fig:app-verification-iters} (Right). 
This suggests that RL improves the quality of generated trajectories and enables the solution agent to solve a broader range of problems. 
However, as shown in Figure~\ref{fig:app-verification-violin} (right), the score gap between never-correct and ever-correct problems narrows to a statistically non-significant margin, with 
$\Delta(\mathrm{wrong}-\mathrm{correct}) = +0.056$ and $p = 0.448$. 
This indicates that although RL strengthens the generation side of TMAS, the verification agent's discriminative ability at the model's new capability frontier remains a limiting factor.

These findings highlight an important limitation of TMAS. 
TMAS improves reasoning by coordinating solution generation, verification, summarization, experience exploitation, and guideline-level exploration. 
However, the overall effectiveness of this synergy is still closely tied to the quality of the verification signal. 
When the solution agent and verification agent approach a shared capability boundary, the verification feedback may become less discriminative, and the downstream experience bank or guideline bank may consequently accumulate less reliable signals. 
In such cases, simply increasing the number of iterations or rollouts may offer limited gains, as the refinement process can be constrained by the reliability of its own feedback loop.

This analysis also points to a valuable direction for future improvement. 
While our current RL training focuses on improving solution generation, experience utilization, and novel strategy exploration, future work could further enhance TMAS by incorporating verification-oriented training. 
Potential directions include training the verification agent with process-level error localization, rewarding the identification of invalid proof steps, calibrating verification scores against ground-truth correctness, or employing stronger and more specialized verification models. 
These extensions may help the verification capability better keep pace with the generation capability, thereby providing more reliable feedback as the system tackles increasingly difficult reasoning problems. 
Overall, strengthening the verification mechanism represents a promising path toward further improving multi-agent synergy in TMAS.

\section{Case Study}
\label{sec:c Case Study}
\label{app:case-study}

\definecolor{softred}{RGB}{252,238,238}
\definecolor{softgreen}{RGB}{235,248,239}
\definecolor{softblue}{RGB}{236,244,252}
\definecolor{softgray}{RGB}{247,247,247}
\definecolor{darkred}{RGB}{155,40,40}
\definecolor{darkgreen}{RGB}{35,120,65}
\definecolor{darkblue}{RGB}{45,85,135}

\tcbset{
  casebox/.style={
    enhanced,
    boxrule=0.5pt,
    arc=2mm,
    left=1.5mm,
    right=1.5mm,
    top=1mm,
    bottom=1mm,
    colframe=black!15,
    fonttitle=\bfseries\small,
    coltitle=black,
  }
}

We present a representative case study to illustrate why the proposed TMAS framework can improve through iterative test-time interaction. The example is Problem 720 in HLE-Math-100, a combinatorics problem that asks for the number of tilings of a $2\times4$ board using $2\times1$, $2\times2$, and $2\times4$ tiles, whose standard answer is $T_4 = 12$.
This case reveals a clear failure mode: without the experience bank, the model repeatedly assumes that the $2\times1$ rectangular tile can only be placed vertically, leading to a persistent wrong answer of $6$. In contrast, TMAS stores and reuses a verified correction signal, allowing the model to eventually solve the problem reliably.

\paragraph{Problem statement.}
Let $T_n$ denote the number of ways to tile a $2\times n$ board using the following tiles:
(i) a $2\times1$ rectangular tile, (ii) a $2\times2$ square tile, and (iii) a $2\times4$ rectangular tile. Compute $T_4$.
\begin{figure*}[t]
\centering
\begin{minipage}[t]{0.48\textwidth}
\begin{tcolorbox}[casebox, colback=softred, title={Wrong solution pattern: vertical-only assumption}, equal height group=mygroup]
\small
The no-experience baseline repeatedly reasons as follows:
\begin{enumerate}[leftmargin=*, itemsep=1pt, topsep=2pt]
    \item A $2\times1$ tile is treated as covering exactly one full column.
    \item A $2\times2$ block is therefore assumed to have only two tilings:
    \[ T(2)=2. \]
    \item The model derives
    \[ T(n)=T(n-1)+T(n-2)+T(n-4). \]
    \item Therefore,
    \[ T(4)=T(3)+T(2)+T(0)=3+2+1=6. \]
\end{enumerate}

\textbf{Diagnostic error.}
The solution explicitly or implicitly rules out horizontal placements of the $2\times1$ tile. By iteration 19, the no-experience baseline even states that horizontal placement is invalid, thereby reinforcing rather than correcting the original mistake.

\[ \boxed{6}\quad\text{\color{darkred}{wrong}} \]
\end{tcolorbox}
\end{minipage}
\hfill
\begin{minipage}[t]{0.48\textwidth}
\begin{tcolorbox}[casebox, colback=softgreen, title={Correct solution pattern: rotation-aware tiling}, equal height group=mygroup]
\small
TMAS eventually reasons as follows:
\begin{enumerate}[leftmargin=*, itemsep=1pt, topsep=2pt]
    \item A $2\times1$ rectangular tile can be placed either vertically or horizontally.
    \item Hence a $2\times2$ block has three tilings:
    \[ T(2)=3, \]
    namely two vertical $2\times1$ tiles, one $2\times2$ square tile, or two horizontal $2\times1$ tiles.
    \item The correct recurrence is
    \[ T(n)=T(n-1)+2T(n-2)+T(n-4). \]
    \item Therefore,
    \[ T(4)=T(3)+2T(2)+T(0)=5+6+1=12. \]
\end{enumerate}

\textbf{Key correction.}
The model explicitly identifies the prior error: the wrong solutions undercount because they assume $T(2)=2$ and ignore the horizontal-pair tiling.

\[ \boxed{12}\quad\text{\color{darkgreen}{correct}} \]
\end{tcolorbox}
\end{minipage}
\caption{Comparison of wrong solution pattern and correct solution pattern.}
\label{fig:case-study-side-by-side}
\end{figure*}

\paragraph{How the experience bank enables correction.}
At iteration 5, most TMAS rollouts still output the wrong answer $6$ (see Figure~\ref{fig:case-study-side-by-side} left). However, one unaided rollout independently discovers that the $2\times1$ tile can be placed horizontally and obtains the correct answer $12$ (see Figure~\ref{fig:case-study-side-by-side} right). The experience extraction agent distills this successful reasoning into reusable entries, including the verified base case $T(2)=3$ and an explicit warning against the vertical-only assumption (see Table~\ref{tab:case-study-experience}).

\begin{table}[h]
\centering
\small
\caption{Key entries extracted into the TMAS experience bank. These entries transform one correct rollout into reusable problem-specific knowledge for later rollouts.}
\label{tab:case-study-experience}
\begin{tabular}{p{0.19\linewidth}p{0.73\linewidth}}
\toprule
\textbf{Entry type} & \textbf{Content} \\
\midrule
Verified anchor 
& $T(2)=3$, verified by enumeration: two vertical $2\times1$ tiles, one $2\times2$ square tile, and two horizontal $2\times1$ tiles stacked vertically to cover a $2\times2$ area. \\
\addlinespace
Verified anchor 
& $T(3)=5$, obtained by enumerating the valid extensions of the $T(2)$ configurations. \\
\addlinespace
Structural rule 
& The correct recurrence is $T(n)=T(n-1)+2T(n-2)+T(n-4)$, where the coefficient $2$ accounts for the two distinct ways to cover a $2$-column block. \\
\addlinespace
Avoidance heuristic 
& Avoid assuming that the $2\times1$ tile can only be placed vertically. This mistake gives $T(2)=2$ instead of $T(2)=3$ and undercounts $T_4$ as $6$ instead of $12$. \\
\bottomrule
\end{tabular}
\end{table}

\paragraph{Transition dynamics.}
The correction does not appear immediately in every rollout. Instead, the improvement emerges gradually (see Table~\ref{tab:case-study-rollouts}). In early iterations, correct rollouts are sparse. After the experience bank begins to expose later rollouts to the verified correction, the number of correct rollouts increases. By iteration 10, more rollouts begin producing $12$, and from iteration 11 onward, the TMAS solution remains nearly correct. The remaining failure is caused by output truncation, where one rollout repeatedly exceeds the maximum length limit and is cut off before giving a valid final answer, rather than by a recurrence of the original reasoning error. 

\begin{table}[h]
\centering
\small
\caption{Selected rollout-level evidence. TMAS gradually increases the number of correct rollouts, whereas the no-experience baseline remains at zero correct rollouts throughout the sampled iterations.}
\label{tab:case-study-rollouts}
\begin{tabular}{c c c}
\toprule
\textbf{Iteration} & \textbf{TMAS correct rollouts / 8} & \textbf{No-experience correct rollouts / 8} \\
\midrule
0  & 0/8 & 0/8 \\
1  & 0/8 & 0/8 \\
5  & 1/8 & 0/8 \\
10 & 4/8 & 0/8 \\
11 & 7/8 & 0/8 \\
15 & 7/8 & 0/8 \\
19 & 7/8 & 0/8 \\
\bottomrule
\end{tabular}
\end{table}


\paragraph{Takeaway.}
This case demonstrates the functional role of the memory bank in TMAS. A single correct rollout is not merely treated as an isolated success; it is converted into persistent, reusable knowledge. The resulting anchor and avoidance heuristic correct a systematic reasoning error, increase the frequency of correct rollouts, and eventually make the corrected solution robust.

\section{Prompt Templates}
\subsection{Prompt Templates for TMAS}
\label{sec:d.1 Prompt Templates for TMAS}
In this section, we comprehensively detail the prompts employed within the TMAS framework. These encompass the system prompt, proof generation prompt, verification prompt, and refinement generation prompt, alongside the integration of experience context and guideline constraints.

\begin{promptbox}{System Prompt}
You are a helpful assistant. To answer the user's question, you first think about the reasoning process and then provide the user with the answer. The reasoning process is enclosed within <think> </think> tags, i.e., <think> reasoning process here </think> answer here.
\end{promptbox}

\begin{promptbox}{Proof Generation Prompt}
Your task is to solve a given problem. The problem may ask you to prove a statement, or ask for an answer. If finding an answer is required, you must come up with the answer, and your final solution must also be a rigorous proof of that answer being valid.

**Goal:** Your objective is to produce a solution that is **exceptionally comprehensive, strictly logical, and easy to follow**.

**Internal Quality Control Standards:**
Before generating your final response, you must internally verify your reasoning against the following strict criteria (though you do not need to output this verification):
1.  **Completeness:** The solution must cover all cases and steps. If minor details are omitted, it is considered imperfect.
2.  **Rigour:** Fatal errors or severe omissions are unacceptable.
3.  **Self-Containment:** Referencing external papers/theorems is allowed **IF AND ONLY IF** you also present a valid proof or clear derivation of the referenced argument. Merely citing a result without showing why it applies or how it works is considered a failure.

**Process:**
1.  Reason carefully about how to solve the problem.
2.  Draft your solution mentally or in your scratchpad.
3.  **Refine your solution** by fixing any potential logical gaps, ambiguity, or "hand-waving" arguments until it meets the highest standard of mathematical proof.
4.  Present *only* your best, finalized version.

**Output Format:**
Your response should follow this exact markdown format:

## Solution
...
// Your final, rigorous solution to the problem here. Ensure all steps are explicitly shown and justified.

---
Here is your task input:
## Problem
{question}
\end{promptbox}

\begin{promptbox}{Verification Prompt}
## Instruction
Your task is to evaluate the quality of a solution to a problem. The problem may ask for a proof of statement, or ask for an answer. If finding an answer is required, the solution should present the answer, and it should also be a rigorous proof of that answer being valid.

Please evaluate the solution and score it according to the following criteria:
- If the solution is completely correct, with all steps executed properly and clearly demonstrated, then the score is 1
- If the solution is generally correct, but with some details omitted or minor errors, then the score is 0.5
- If the solution does not actually address the required problem, contains fatal errors, or has severe omissions, then the score is 0
- Additionally, referencing anything from any paper does not save the need to prove the reference. It's okay IF AND ONLY IF the solution also presents a valid proof of the reference argument(s); otherwise, if the solution omits the proof or if the proof provided is not completely correct, the solution should be scored according to the criteria above, and definitely not with a score of 1.

Please carefully reason out and analyze the quality of the solution below, and in your final response present a detailed evaluation of the solution's quality followed by your score. Therefore, your response should be in the following format:

Here is my evaluation of the solution:
...
// Your evaluation here. You are required to present in detail the key steps of the solution or the steps for which you had doubts regarding their correctness, and explicitly analyze whether each step is accurate: for correct steps, explain why you initially doubted their correctness and why they are indeed correct; for erroneous steps, explain the reason for the error and the impact of that error on the solution.

Based on my evaluation, the final overall score should be:
\boxed{...}
// where ... should be the final overall score (0, 0.5, or 1, and nothing else) based on the above criteria.

---
Here is your task input:
## Problem
{question}
## Solution
{proof}
\end{promptbox}

\begin{promptbox}{Refine Generation Prompt}
Your task is to solve a given problem. The problem may ask you to prove a statement, or ask for an answer. If finding an answer is required, you must come up with the answer, and your final solution must also be a rigorous proof of that answer being valid.

**Context & Objective:**
Previous solution attempts have been made on this problem. Each attempt comes with a verification summary that identifies specific errors, gaps, or disputed steps. Your primary goal is to **carefully analyze these previous attempts and their feedback**, then produce a **new, corrected solution** that directly addresses the identified flaws.

**Internal Quality Control Standards:**
Before generating your final response, you must internally verify your reasoning against the following strict criteria:
1.  **Error Correction:** You must explicitly address the flaws pointed out in the verification summaries. Do NOT repeat logic that has already been identified as incorrect.
2.  **Completeness:** The solution must cover all cases and steps. If minor details are omitted, it is considered imperfect.
3.  **Rigour:** Fatal errors or severe omissions are unacceptable.
4.  **Self-Containment:** Referencing external papers/theorems is allowed **IF AND ONLY IF** you also present a valid proof or clear derivation of the referenced argument.

**Process:**
1.  Read the **Problem** carefully.
2.  Study each **Previous Attempt** and its **Verification Summary**. Identify exactly what went wrong, what was incomplete, and what (if anything) was correct.
3.  Reason about how to fix the specific issues while retaining any correct sub-results from previous attempts.
4.  Draft your refined solution, ensuring it does not repeat the confirmed errors.
5.  Present *only* your best, finalized, and fully corrected version.

**Output Format:**
Your response should follow this exact markdown format:

## Solution
...
// Your final, rigorous, and corrected solution to the problem here. Ensure all steps are explicitly shown and justified.

---
Here is your task input:
## Problem
{question}


\end{promptbox}

\begin{promptbox}{Experience Context Appended After the Refine Generation Prompt}
{Refine Generation Prompt mentioned above}

[Accumulated Experience \& Constraints]

### Verified Anchors (Proven Facts)
The following conclusions have already been rigorously verified.
**Action:** Use them directly as premises. Do NOT re-prove them.
{anchors_list}

### Strategic Heuristics (Methods \& Pitfalls)
The following are verified strategies and identified pitfalls from previous attempts.
**Action:** Prioritize these methods and strictly avoid the errors mentioned.
{heuristics_list}
\end{promptbox}


\begin{promptbox}{Guideline Constraint Appended After the Refine Generation Prompt}
{Refine Generation Prompt mentioned above}

[Exploration Directive --- Read Carefully Before Generating]

## What Are These Guidelines?
The entries below are a log of **high-level solution strategies** that have already been attempted on this exact problem in previous iterations. Each entry describes the broad mathematical framework, key structural insight, and angle of attack used --- for example: which technique was applied (induction, generating functions, algebraic manipulation, geometric transformation, probabilistic argument, etc.), what the central idea was, and why it ultimately failed or fell short.

## Why You Must Choose a Different Approach
This system runs many iterations to explore the solution space. Repeating a strategy that has already been tried wastes an entire iteration and produces no new information. The verification system has already evaluated these approaches and found them insufficient to produce a fully correct solution. Attempting the same strategy again --- even with minor tweaks in notation or presentation --- will almost certainly lead to the same failure modes.

To help us effectively explore the solution space, please try to avoid repeating the exact same paths that have already proven unsuccessful.

You are encouraged to explore a new direction. This could mean adopting a different mathematical framework entirely, or it could mean using a similar foundational approach while making distinctly different choices in your intermediate steps, structural manipulations, or angles of attack to bypass the previous pitfalls.

## Already-Attempted Strategies (DO NOT REPEAT ANY OF THESE):
{tried_list}
\end{promptbox}

\begin{promptbox}{Experience Evolution Template}
You are the **Curator of a Mathematical Experience Bank**.

## What is the Experience Bank?
The Experience Bank stores **low-level, concrete, directly actionable knowledge** extracted from actual solution attempts. This is the fine-grained, technical residue of working through the mathematics itself -- not high-level strategy (that belongs in the Guideline Bank).
The bank contains two categories of entries:
---
### Category 1: Verified Anchors
A Verified Anchor is a **non-trivial intermediate result** that has been confirmed correct by verifiers and is worth preserving so that future solvers can build on it without re-deriving it. An anchor must meet **all three** of the following criteria:
1. **Non-trivial**: It must involve meaningful mathematical work -- a derivation, a transformation, a non-obvious equivalence, or a structural observation. Trivial arithmetic evaluations (e.g., "2025 == 2 mod 7") do NOT qualify unless the congruence itself is the key insight that unlocks a deeper argument.
2. **Reusable**: It must be a stepping stone -- something a future solver can directly cite and proceed from, without needing to redo the work. Prefer results that establish structure over results that are dead ends.
3. **Verifier-backed**: It must be explicitly confirmed correct by the verification summary. If verifiers are split on a step, do not add it as an Anchor.
Verified Anchors fall into the following sub-types (use these to guide what you extract):
- **Structural Reduction**: A transformation that rewrites the problem or a sub-problem into a simpler or more tractable form.
  - Example: "The substitution $u = x - 1/x$ reduces the integral to $\int \\frac{{du}}{{u^2+2}}$, which is a standard form."
- **Algebraic Equivalence**: A non-obvious algebraic identity or simplification that holds specifically for this problem's constraints.
  - Example: "Under the constraint $a + b + c = 1$, the expression $a^2 + b^2 + c^2 - ab - bc - ca$ simplifies to $1 - 3(ab + bc + ca)$."
- **Logical Implication/Domain Constraint**: A deduction that narrows the solution space or establishes a necessary condition -- particularly useful for eliminating cases.
  - Example: "From the parity argument, the LHS is always even, so $n$ must be even. This eliminates all odd $n$ from consideration."
- **Correct Application of a Theorem or Identity**: A confirmed correct instantiation of a known result (e.g., AM-GM, Cauchy-Schwarz, Vieta's, Fermat's Little Theorem) in the specific context of this problem, including verification that the applicability conditions are met.
  - Example: "AM-GM applies here because all terms are positive under the constraint $x, y > 0$, giving $x/y + y/x \geq 2$. The equality condition $x = y$ is achievable."
- **Boundary/Extremal Result**: A confirmed extremal value, equality case, or boundary behavior that characterizes the solution.
  - Example: "The maximum of $f(x) = x(1-x)$ on $[0,1]$ is $1/4$, achieved at $x = 1/2$."
---
### Category 2: Error Avoidance Heuristics
An Error Avoidance Heuristic records a **specific, concrete pitfall** that was encountered in a solution attempt and confirmed as an error by verifiers. Its purpose is to warn future solvers away from traps that look plausible but lead to failure.
A good heuristic must:
- Identify the **specific step or reasoning pattern** that failed (not just "the solution was wrong")
- Explain **why** it fails in the context of this problem
- Be specific enough that a solver can recognize and avoid it
Examples:
- "Squaring both sides of the inequality at step 3 introduces extraneous solutions when $x < 0$. Always case-split on the sign of $x$ before squaring."
- "The naive application of AM-GM fails here because the equality condition requires $a = b = c$, which is incompatible with the constraint $a + b + c = 0$. Do not use AM-GM without verifying the equality condition is reachable."
- "Treating the recurrence as linear and applying the characteristic equation ignores the non-linear term at step 5. The linearization is invalid beyond first order."
---
## What does NOT belong in the Experience Bank:
- High-level strategic directions (e.g., "try induction", "use a generating function approach") -- those belong in the Guideline Bank.
- Trivial arithmetic facts with no structural significance (e.g., "2025 = 45^2", "7 * 289 = 2023") -- these provide no leverage to a future solver.
- Vague general advice not tied to specific steps in the solution (e.g., "be careful with signs", "check boundary cases").
- Any result from a low-scoring rollout that has not been explicitly validated by the verification summary.
---
## Your Task
You are given all solution attempts (rollouts) from the current iteration, each paired with a verification summary. Your task is to **UPDATE** the existing Experience Bank by extracting new knowledge from these rollouts.
## Operations
1. **ADD**: Extract new Verified Anchors or Error Avoidance Heuristics that are not already covered by the existing bank. Prioritize insights confirmed consistently across multiple rollouts.
2. **KEEP**: Retain all existing entries that remain valid and are not contradicted by the new rollouts.
3. **REFINE**: If a new rollout provides a more precise version of an existing entry, rewrite it to be clearer. Only merge entries that say the exact same thing about the exact same step.
4. **DELETE**: Remove entries explicitly revealed as incorrect by the verification summary. Remove entries that become fully subsumed after refinement.
## Quantity Guideline
Aim for **20-35 entries** in total across both categories. Do NOT aggressively compress -- fine-grained, specific entries are more useful than over-generalized ones. Only merge entries that are truly redundant.
## Output Requirements
Output the **FULL updated bank** in JSON format (replacing the old bank entirely):
```json
{{
    "verified_anchors": [
        "<sub-type>: <concrete verified intermediate result with enough context to be self-contained>",
        ...
    ],
    "error_avoidance_heuristics": [
        "Avoid: <specific pitfall, why it fails in this problem, and what to do instead>",
        ...
    ],
    "changes_log": "Brief summary of what was added, refined, or removed and why."
}}
```

When writing a Verified Anchor, prefix it with its sub-type (e.g., `Structural Reduction:`, `Algebraic Equivalence:`, `Logical Implication:`, `Theorem Application:`, `Boundary Result:`).
## Task Input
### Problem:
{question}
### Existing Bank (Current State):
{existing_experiences}
### Current Iteration Rollouts:
{rollouts}
\end{promptbox}

\begin{promptbox}{Guideline Update Template}
You are a **Strategic Advisor** managing the Guideline Bank for an iterative mathematical problem-solving system.
## What is the Guideline Bank?
The Guideline Bank is a **log of high-level solution strategies** that have been attempted so far on this problem. Its purpose is twofold:
1. **Memory of exploration**: It records which broad strategic directions have already been tried, so the solver does not waste computation repeating the same approach.
2. **Diversity enforcement**: When the solver is about to generate a new solution, it will be shown this bank and instructed to pursue a direction that is **fundamentally different** from everything listed here. The bank therefore acts as the primary mechanism for controlling exploration -- the richer and more precise this log is, the better the solver can navigate away from exhausted directions.
## What is a "Guideline" entry?
A guideline entry describes the **high-level conceptual approach and proof strategy** of a solution attempt -- not the low-level computation details (those belong in the Experience Bank). A good guideline entry answers:
- What is the overall mathematical framework or technique being used? (e.g., induction, generating functions, probabilistic method, algebraic manipulation, geometric transformation, etc.)
- What is the key structural insight or angle of attack? (e.g., "reduce to a fixed-point problem", "rewrite as a telescoping sum", "embed in a higher-dimensional space")
- If relevant: what makes this attempt distinct from others already in the bank?
A guideline entry should be **concise (2-4 sentences)** but **precise enough** that a future solver can clearly understand what approach it refers to and consciously choose a different one.
**Examples of good guideline entries:**
- "Attempted a direct induction on n. The inductive step tried to bound the sum by splitting into two halves, but the resulting inequality was too weak to close. Approach: elementary induction with splitting argument."
- "Used generating functions: encoded the recurrence as a rational function and attempted partial fraction decomposition to extract closed-form coefficients. The poles were algebraically intractable due to an irreducible quartic denominator."
- "Geometric approach: interpreted the inequality as a statement about convexity of a curve and attempted to apply Jensen's inequality. Failed because the function was not convex on the full domain."
**What does NOT belong in the Guideline Bank:**
- Low-level computation details or specific algebraic steps (those belong in the Experience Bank).
- Vague entries like "tried algebra" or "used calculus" -- entries must be specific enough to be distinguishable.
## Your Task
You are given all solution attempts (rollouts) from the current iteration. For each rollout, extract a guideline entry describing its high-level strategy. Then update the existing Guideline Bank by adding the new entries.
**Important: The Guideline Bank is append-only (only grows, never shrinks).** Do not remove or modify existing entries -- only add new ones. If a new rollout's strategy is essentially identical to an existing entry, do not add a duplicate; instead, note it in the changes log.
## Output Requirements
Output the **complete updated bank** (all existing entries + new entries) in JSON format:
```json
{{
    "updated_guidelines": [
        "Guideline 1: <high-level strategy description>",
        "Guideline 2: ...",
        ...
    ],
    "changes_log": "Brief explanation of which new strategies were added and which rollouts were considered duplicates of existing entries."
}}
```
## Task Input
### Problem:
{question}
### Existing Guideline Bank:
{existing_guidelines}
### Current Iteration Rollouts:
{rollouts}
\end{promptbox}

\subsection{Prompt Template for Novel Exploration Reward}
\label{sec:d.2 Prompt Template for Novel Exploration Reward}
\begin{promptbox}{Guideline Judge in RL Training}
You are an expert mathematical strategy analyst. Your task is to determine whether a student's solution follows the exploration directive given to them -- specifically, whether they repeated any of the previously-attempted strategies that they were explicitly instructed to avoid.

### Classification Rules:

**Label 0 (Violated -- strategy was repeated):**
- The student's solution uses a high-level approach, mathematical framework, or angle of attack that is essentially the same as one of the already-attempted strategies listed in the directive.
- Minor surface-level differences in notation or presentation do NOT count as a distinct strategy.

**Label 1 (Compliant -- genuinely new strategy):**
- The student's solution adopts a fundamentally different approach from all listed strategies.
- It may share some basic tools (e.g., both use algebra) but takes a clearly different structural route, key insight, or angle of attack.

**Label -1 (Unable to Judge):**
- The solution is too short, too vague, or too incomplete to determine its high-level strategy.

### Output Format:
Respond strictly in a JSON code block. Example:
```json
{"identified_strategy": "...", "matched_guideline": null, "reasoning": "...", "label": 1}
```
Do NOT output any text outside the code block.
"""

GUIDELINE_JUDGE_USER_TEMPLATE = """
<problem>
{problem}
</problem>

<already_attempted_strategies>
{tried_list}
</already_attempted_strategies>

<student_solution>
{student_solution}
</student_solution>

[Task]
Determine whether the student's solution repeats any of the already-attempted strategies.

1. **Identify** the high-level strategy used in the student's solution (mathematical framework, key structural insight, angle of attack).
2. **Compare** it against each entry in <already_attempted_strategies>.
3. **Classify**:
   - **1**: The student's strategy is genuinely different from ALL listed strategies.
   - **0**: The student's strategy is essentially the same as at least one listed strategy.
   - **-1**: Cannot determine the student's strategy (solution too vague/incomplete).

Respond in a JSON code block:
```json
{{
    "identified_strategy": "Brief description of the strategy used in the student's solution...",
    "matched_guideline": "The guideline entry it matches, or null if none matched",
    "reasoning": "Explanation of why you assigned this label...",
    "label": 1
}}
```
\end{promptbox}

\end{document}